\documentclass[conference]{IEEEtran}
\IEEEoverridecommandlockouts
\usepackage{stfloats}
\usepackage{float}
\usepackage{cite}
\usepackage{amsmath,amssymb,amsfonts}
\usepackage{algorithmic}
\usepackage{graphicx}
\usepackage{textcomp}
\usepackage{xcolor}
\usepackage{booktabs}
\usepackage{multirow}
\usepackage{subcaption}
\usepackage{marvosym}
\usepackage{makecell}

\usepackage[colorlinks,
linkcolor=blue,
anchorcolor=blue,
citecolor=blue]{hyperref}
\def\BibTeX{{\rm B\kern-.05em{\sc i\kern-.025em b}\kern-.08em
    T\kern-.1667em\lower.7ex\hbox{E}\kern-.125emX}}
\makeatletter
\newcommand{\linebreakand}{\end{@IEEEauthorhalign}\hfill\mbox{}\par\mbox{}\hfill\begin{@IEEEauthorhalign}}
\makeatother

\begin{document}

\title{MTS-UNMixers: Multivariate Time Series Forecasting via Channel-Time Dual Unmixing
\thanks{This work was supported by Dalian Science and Technology InnovationFund Project 2022JJ11CG002. The corresponding author is Hongyu Wang.}
}
\author{\IEEEauthorblockN{1\textsuperscript{st} Xuanbing Zhu}
\IEEEauthorblockA{\textit{School of Information and} \\ \textit{Communication Engineering} \\
\textit{Dalian University of Technology}\\
DaLian, China  \\
zhuxuanbing@mail.dlut.edu.cn}
\and
\IEEEauthorblockN{2\textsuperscript{nd} Dunbin Shen}
\IEEEauthorblockA{\textit{School of Information and} \\ \textit{Communication Engineering} \\
\textit{Dalian University of Technology}\\
DaLian, China\\
sdb\_2012@163.com}
\and
\IEEEauthorblockN{3\textsuperscript{rd} Zhongwen Rao}
\IEEEauthorblockA{\textit{ Huawei Noah's Ark Lab} \\
ShenZhen, China \\
raozhongwen@huawei.com}
\linebreakand
\IEEEauthorblockN{4\textsuperscript{th} Huiyi Ma}
\IEEEauthorblockA{\textit{School of Information and} \\ \textit{Communication Engineering} \\
\textit{Dalian University of Technology}\\
DaLian, China   \\
mahuiyi@mail.dlut.edu.cn}
\and
\IEEEauthorblockN{5\textsuperscript{th} Yingguang Hao}
\IEEEauthorblockA{\textit{School of Information and} \\ \textit{Communication Engineering} \\
\textit{Dalian University of Technology}\\
DaLian, China   \\
yghao@dlut.edu.cn}
\and
\IEEEauthorblockN{6\textsuperscript{th} Hongyu Wang\textsuperscript{\Letter}}
\IEEEauthorblockA{\textit{School of Information and} \\ \textit{Communication Engineering} \\
	\textit{Dalian University of Technology}\\
DaLian, China  \\
whyu@dlut.edu.cn}
}

\maketitle
\begin{abstract}
Multivariate time series data provide a robust framework for future predictions by leveraging information across multiple dimensions, ensuring broad applicability in practical scenarios. However, their high dimensionality and mixing patterns pose significant challenges in establishing an interpretable and explicit mapping between historical and future series, as well as extracting long-range feature dependencies. To address these challenges, we propose a channel-time dual unmixing network for multivariate time series forecasting (named MTS-UNMixer), which decomposes the entire series into critical bases and coefficients across both the time and channel dimensions. This approach establishes a robust sharing mechanism between historical and future series, enabling accurate representation and enhancing physical interpretability. Specifically, MTS-UNMixers represent sequences over time as a mixture of multiple trends and cycles, with the time-correlated representation coefficients shared across both historical and future time periods. In contrast, sequence over channels can be decomposed into multiple tick-wise bases, which characterize the channel correlations and are shared across the whole series. To estimate the shared time-dependent coefficients, a vanilla Mamba network is employed, leveraging its alignment with directional causality. Conversely, a bidirectional Mamba network is utilized to model the shared channel-correlated bases, accommodating noncausal relationships. Experimental results show that MTS-UNMixers significantly outperform existing methods on multiple benchmark datasets. The code is available at \url{https://github.com/ZHU-0108/MTS-UNMixers}.
\end{abstract}

\begin{IEEEkeywords}
	time series forecasting, unmixing, mamba network, shared.
\end{IEEEkeywords}

\section{Introduction}
The time series forecasting aims to provide accurate predictions of future series values by analyzing time-dependent patterns and trends in historical data. It is a core task in data analytics and is widely used in financial markets\cite{chen2012bayesian}, weather forecasting\cite{angryk2020multivariate,schultz2021can}, electric power forecasting\cite{khan2020towards,zhu2023energy}, and traffic flow estimation\cite{chen2001freeway,cirstea2022towards,ma2014enabling}, among other fields. With the exponential growth of data volumes and significant advancements in computational power, predictive models now face two key challenges. The first challenge is modeling complex nonlinear relationships over time to accurately capture essential features within long-term sequences. The second involves extracting interactions from multivariate data to better understand the dynamic variations in time series and identify latent patterns. To address these challenges, models have to effectively integrate data from multiple sources and also have robust prediction capabilities.

Recently, many deep learning models have been proposed to address the challenges of long series dependence and multivariate modeling in multivariate time series data. To capture long-series features more effectively, Chen et al.\cite{chen2024pathformer} introduced adaptive multiscale modeling with temporal dynamic inputs, which leverages both local and global information on the time axis. Wang et al.\cite{TimeMachine} integrated multiscale time series by incorporating micro-seasonal information and macro-trend data, utilizing complementary forecasting capabilities to improve overall performance. In order to extract the interaction information in multivariate time series, Liu et al.\cite{liu2023itransformer} redesigned the Transformer architecture to utilize the attention mechanism to capture the correlation between variables, and at the same time utilized the feed-forward neural network to extract the temporal features, which effectively enhances the cross-channel correlation extraction. Meanwhile, Li et al.\cite{li2023mtsmixersmultivariatetimeseries} employed a decomposition module to capture inter-channel relationships. Its relatively simple structure offers significant efficiency advantages over more complex attention-based mechanisms.

\begin{figure*}[ht]
	\centering
	\captionsetup[subfigure]{labelformat=empty}
	\begin{minipage}{0.45\linewidth}
		\vspace{3pt}
		\includegraphics[width=\textwidth, trim=0.5cm 0.2cm 0.4cm 0.2cm, clip]{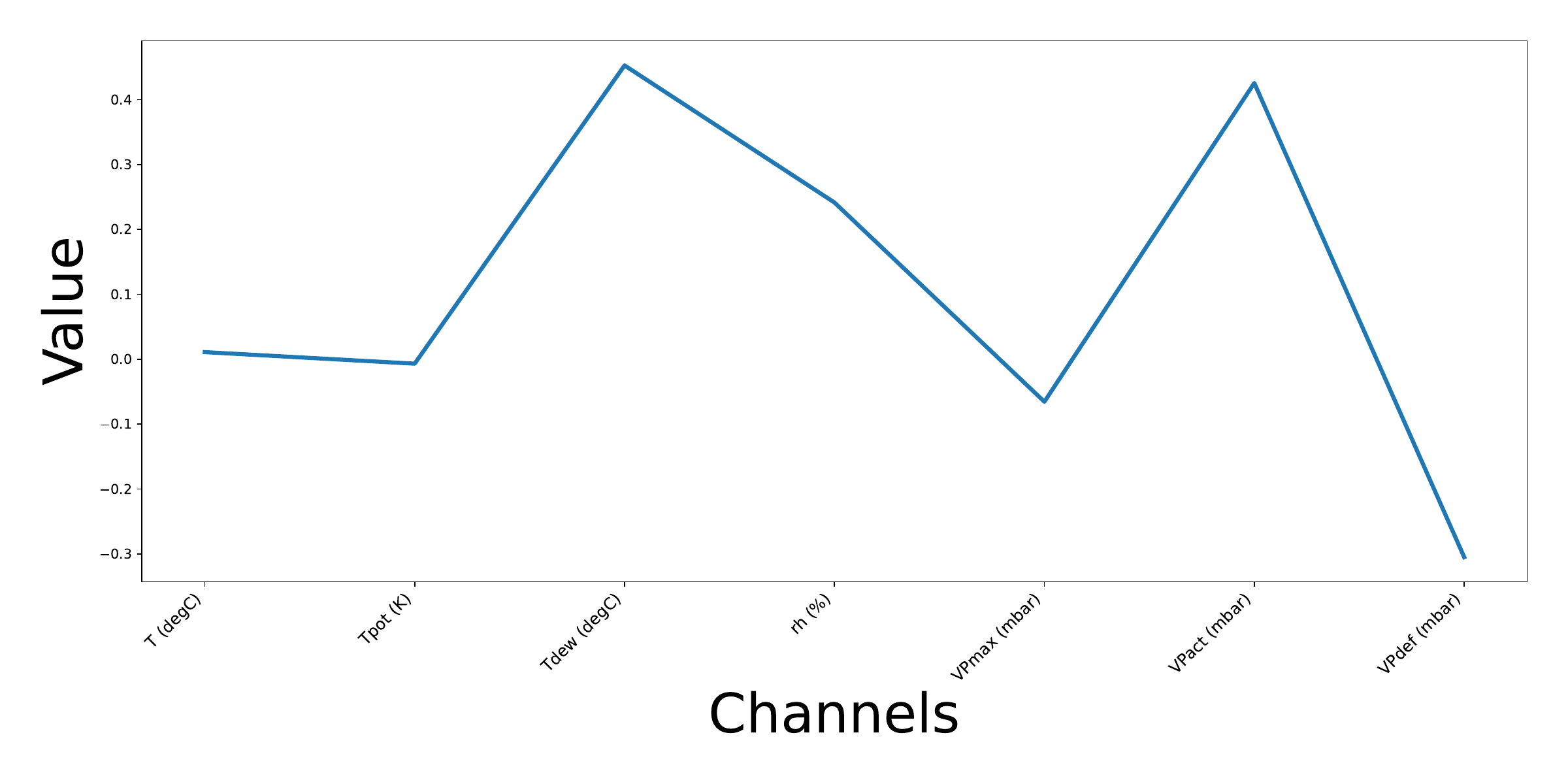}
		\vspace{3pt}
	\end{minipage}
	\begin{minipage}{0.45\linewidth}
		\vspace{3pt}
		\includegraphics[width=\textwidth,  trim=0.5cm 0.2cm 0.4cm 0.2cm, clip]{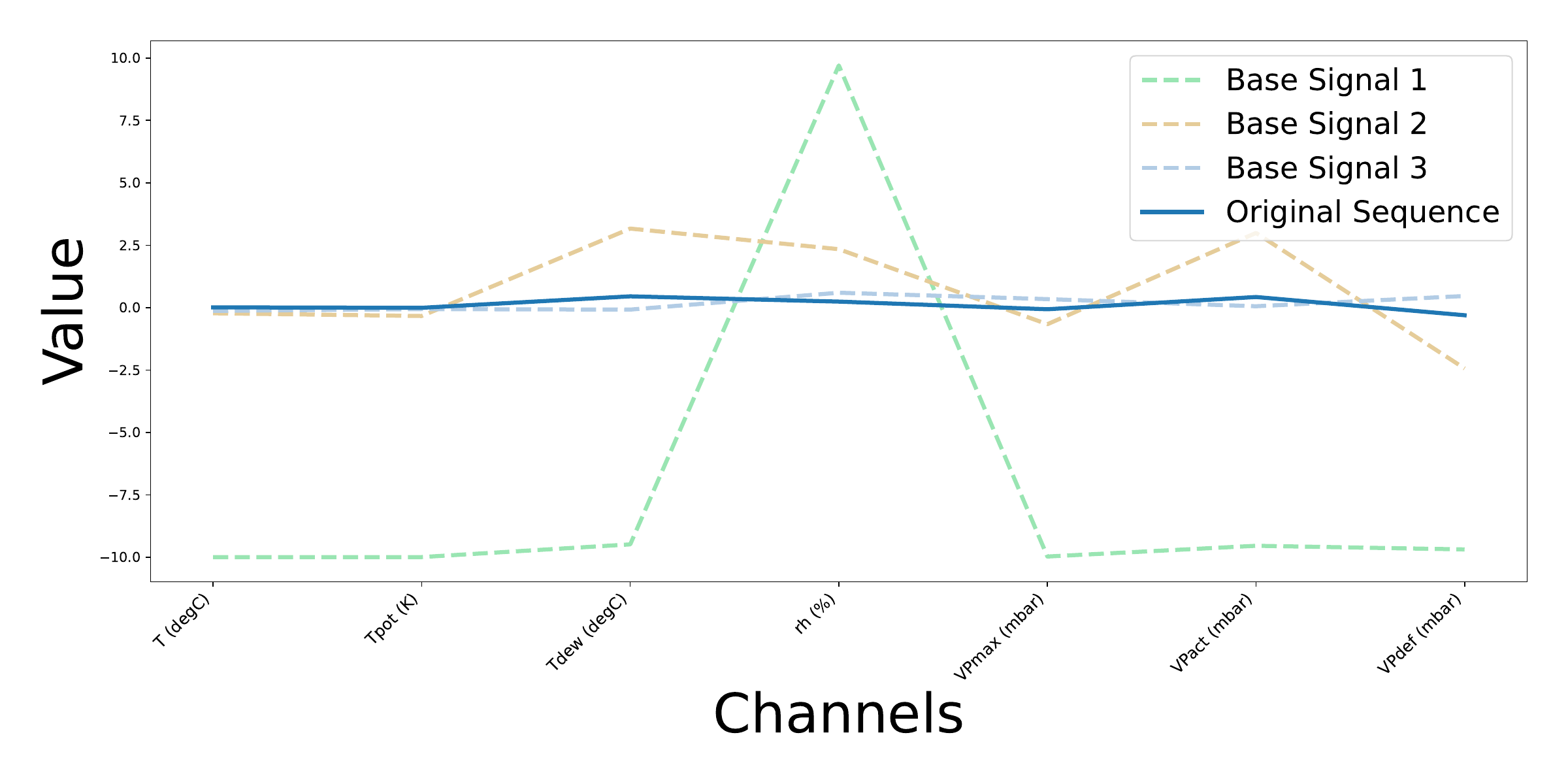}
		\vspace{3pt}
	\end{minipage}
	\vspace{2pt} 
	\centerline{(a) Channel mixing.}
	\begin{minipage}{0.45\linewidth}
		\vspace{3pt}
		\includegraphics[width=\textwidth, trim=0.5cm 0.2cm 0.4cm 0.2cm, clip]{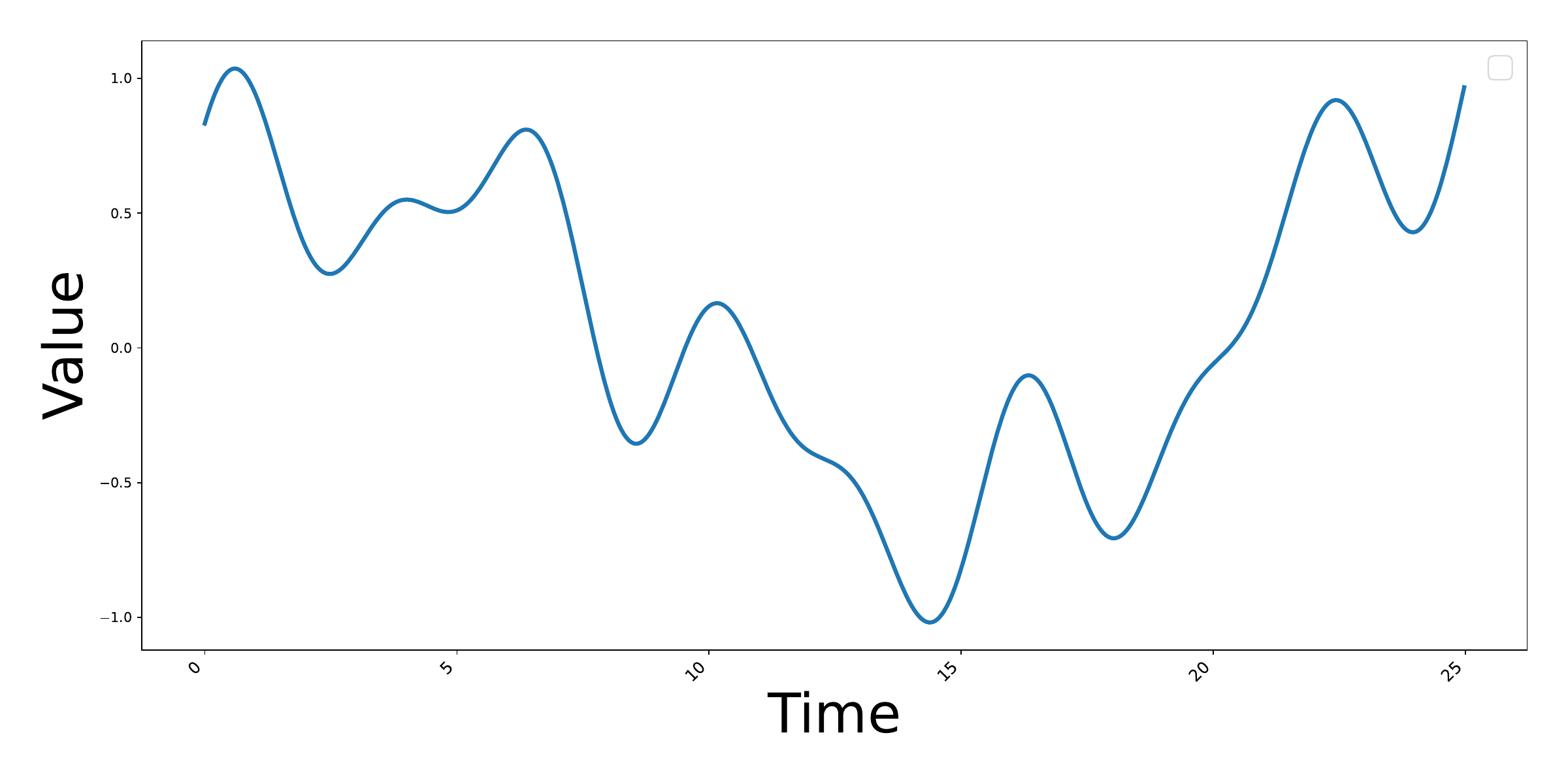}
		\vspace{3pt}
	\end{minipage}
	\begin{minipage}{0.45\linewidth}
		\vspace{3pt}
		\includegraphics[width=\textwidth, trim=0.5cm 0.2cm 0.4cm 0.2cm, clip]{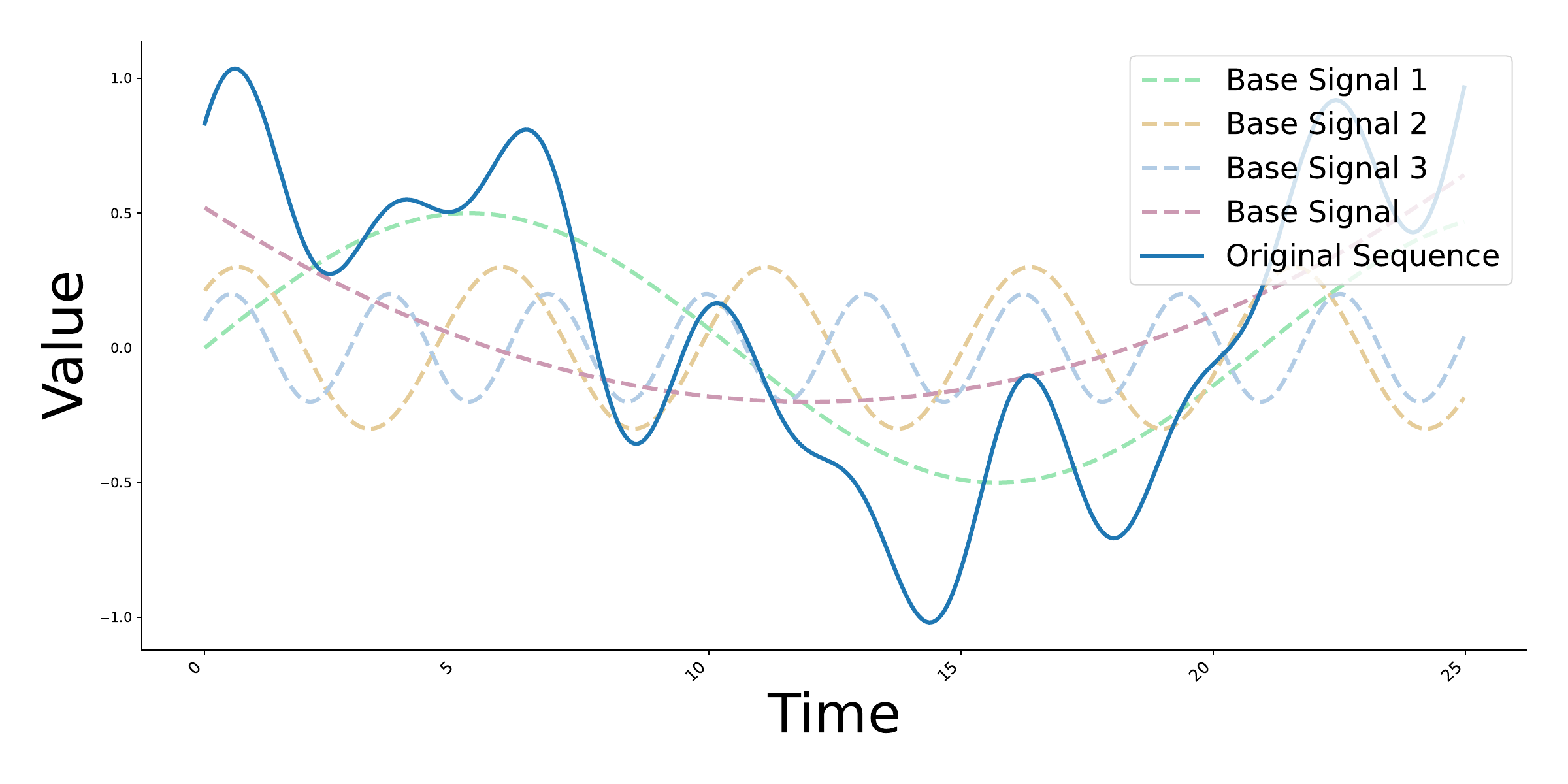}
		\vspace{3pt}
	\end{minipage}
	\vspace{2pt} 
	\centerline{(b) Time mixing.}
	\caption{Mixing problem  (taking weather data as an example). (a) shows the original plot of the seven variables and their principal component composition in the weather dataset. The high correlation between variables suggests that they may be influenced by common external environmental factors. (b) illustrates the time series of a single channel and shows how it can be decomposed into a mixture of different features.}
	\label{introduction1}
\end{figure*}	

Although recent advances have been made in dealing with long series features and multivariate interactions, these models still face limitations. The first challenge is the mixing problem in multivariate time series data, where different features intertwine with each other during the modeling process, making it difficult for the model to investigate individual contributions from each feature. Specifically:
\begin{itemize}
\item  In the channel dimension, multivariables tend to exhibit high correlation, leading to feature redundancy. As shown in the weather dataset in Fig. \ref{introduction1} (a), variables such as temperature, humidity, and cloud cover are all affected by the dominant temperature trend, which results in similar variations between variables. This similarity reflects the natural correlation between weather variables and may lead to feature overlap and blending, which in turn increases computational complexity and amplifies noise during channel blending.
\item In the time dimension, different periods often represent a mixture of patterns, such as cycles and trends, rather than a single feature. For example, in the time series data shown in Fig. \ref{introduction1} (b), we can observe multiple cyclical features (e.g., daily variations or seasonal fluctuations) coexisting with long-term trends. These patterns overlap each other on the time axis, which not only complicates component separation, but also masks localized patterns and interferes with long-term dependencies.
\end{itemize}

The second challenge lies in the limitations of traditional ``black-box" mapping models in modeling the relationship between historical and future sequences. Black-box mappings typically rely on abstract feature representations and often lack the ability to precisely map between sequences. They also does not have an explicit physical meaning and lacks interpretability of the entire sequence characteristics.

To address the above limitations, we propose MTS-UNMixers, which establishes an explicit mapping of sequences through a ummixing mechanism. MTS-UNMixers decouples the historical sequences in time and channel, and extracts the significant components on the time and channel axes.  Specifically, MTS-UNMixers uses a mixture model to represent a single-channel (multi-time) series as a combination of several trend and period components. For a single moment (multi-variate) it can be represented by a number of cardinalities which describe the correlation between the individual variables, called correlation components.The historical time horizon and the future time horizon are treated as a unified whole, allowing for unmixed and explicit mapping by sharing underlying signal and weight information.
In temporal unmixing, we use the Mamba network.  The network effectively extracts the nonlinear causal dependencies in the sequence through the dynamic causal mechanism in the state-space model, by utilising the nonlinear multiplication between the basis signal matrix and the component coefficient matrix.
In channel unmixing, we employ a bidirectional Mamba network since there is no causal relationship between channels, but rather a bidirectional interaction between highly correlated variables. In this way, the model captures the bidirectional correlations and interactions between variables, realizes the decoupling of independent features in multiple channels, and improves the clarity and accuracy of feature extraction.

In summary, the main contributions of this paper are as follows:

\begin{itemize}
	\item To address the problem of reduced model accuracy due to feature overlap and correlation in high-dimensional time series data, we propose a dual-mixing model to decompose the entire series into critical bases and coefficients along both the time and channel dimension. This decomposition can effectively investigate the individual contributions of intrinsic time-dependent and channel-correlated bases, reducing the noise and redundancy.
	\item For effective sharing, we adopt an explicit mapping model. Historical and future sequences are treated as a unified whole, sharing the component coefficient matrix in the channel dimension and the base signal matrix in the time dimension. This improves physical interpretability and prediction reliability while ensuring the conciseness and simplicity of the model.
	\item In the time dimension, we use Mamba to capture the characteristics of long-term temporal dependencies, while channel dimension where no causal relationship exists, we use bi-directional Mamba to effectively capture the bidirectional interactions between variables.
\end{itemize}

The remainder of this article is organized as follows.
First, we provide a brief overview of the relevant research on time series forecasting in Section \ref{Related}. Second, we describe the problem in Section \ref{mixing}. Next, the proposed MTS-UNMixers method is presented in detail in Section \ref{unmxing}. Then, the experiments and corresponding analyses are presented in Section \ref{reslut}. Finally, conclusions are drawn in Section \ref{conclusion}.
\section{Related Work}\label{Related}
\subsection{Deep learning Models for Time Series Forecasting}
In recent years, deep learning has been widely applied in time series forecasting, and the models can be broadly categorized into attention-based and non-attention-based approaches. Attention-based models primarily include the Transformer series, which utilize self-attention mechanisms to capture long-range dependencies and relationships between different time points in time series data. These models are particularly suited for forecasting tasks involving long sequences and high-dimensional data. Transformer models, with self-attention mechanisms, excel in capturing long-range dependencies and inter-timepoint relationships, making them well-suited for long-sequence and high-dimensional forecasting tasks \cite{nietime}, \cite{zhou2022fedformer}, \cite{wu2021autoformer}, \cite{liu2022non} .Nie et al. proposed PatchTST\cite{nietime}, a Transformer-based model design that enhances efficiency and long-term forecasting accuracy for multivariate time series by leveraging time series segmentation and channel independence. Zhou et al. proposed FEDformer\cite{zhou2022fedformer} a method that combines Transformer with seasonal-trend decomposition and frequency enhancement to effectively capture both global structure and detailed features of time series, improving long-term forecasting performance.

Non-attention-based models include recurrent neural network (RNN)-based, convolutional neural network (CNN)-based, mamba-based and multi-layer perceptron (MLP)-based models.These models capture local dependencies, time series characteristics, and multi-scale features in time series data through various modeling approaches. Technically, MLP-based methods apply multi-layer perceptrons along the temporal dimension, offering simplicity and competitive performance, especially with high-dimensional data \cite{TimeMachine}, \cite{ekambaram2023tsmixer}, \cite{zeng2023transformers}. Wang et al. \cite{TimeMachine} proposed TimeMixer, a fully MLP-based architecture that leverages Past-Decomposable-Mixing (PDM) and Future-Multipredictor-Mixing (FMM) modules to disentangle and capture multiscale temporal variations for enhanced time series forecasting. CNN-based models leverage convolutional kernels to effectively capture local patterns over time \cite{liu2022scinet}, \cite{wu2023timesnet}, \cite{luo2024moderntcn}. Liu et al. proposed SCINet\cite{liu2022scinet}, a novel neural network architecture that recursively downsamples, convolves, and interacts to effectively model complex temporal dynamics, achieving significant improvements in forecasting accuracy. RNN-based models utilize recurrent structures to manage temporal dependencies in sequential data \cite{salinas2020deepar}, \cite{zhao2017lstm}. David Salinas et al. proposed the DeepAR method\cite{salinas2020deepar}, which achieves high-precision probabilistic forecasting by training an auto-regressive recurrent network model on a large collection of related time series. 
Recently, Gu and Dao presented Mamba\cite{gu2023mamba}, integrating parameterized matrices and hardware-aware parallel computing, achieving superior performance in multiple tasks. Derived from Mamba, several models have further advanced time series forecasting. Ma et al. proposed FMamba\cite{FMamba}, combining fast-attention with Mamba for efficient temporal and inter-variable dependency modeling. Patro and Agneeswaran introduced SiMBA\cite{SiMBA}, a simplified Mamba-based architecture using EinFFT for channel modeling, outperforming existing SSMs in both image and time series benchmarks. Tang et al.\cite{Tang_CVPR} introduced VMRNN, integrating Vision Mamba blocks with LSTM for enhanced spatiotemporal forecasting with a smaller model size. Ahamed and Cheng proposed TimeMachine\cite{timemachine}, a quadruple-Mamba architecture achieving superior accuracy, scalability, and memory efficiency in long-term time series forecasting. However, despite the promising progress, the full potential of the Mamba model in more complex and challenging prediction tasks has not been fully utilised, indicating a need for further exploration and refinement.
\subsection{State space models}
State Space Models (SSMs) \cite{rangapuram2018deep, lin2021ssdnet, gu2022efficiently, gu2023mamba} use intermediate state variables to achieve sequence-to-sequence mapping, enabling efficient handling of long sequences. The core concept of SSMs is to capture complex dynamic features through state variables, addressing the computational and memory bottlenecks encountered in traditional long-sequence modeling. However, early SSM models often required high-dimensional matrix operations, leading to substantial computational and memory demands. Rangapuram et al. proposed Deep State Space Models (DSSM)\cite{rangapuram2018deep}, integrating latent states with deep neural networks to better model complex dynamics. Lin et al. introduced SSDNet\cite{lin2021ssdnet}, which combines Transformer architecture with SSMs for efficient temporal pattern learning and direct parameter estimation, avoiding Kalman filters. 

To overcome these limitations, structured state-space models like S4 \cite{gu2022efficiently} introduced innovative improvements. S4 employs low-rank corrections to regulate certain model parameters, ensuring stable diagonalization and reducing the SSM framework to a well-studied Cauchy kernel. This transformation not only significantly reduces computational costs but also maintains performance while drastically lowering memory requirements, making SSM more practical for long-sequence modeling.

Building on this foundation, Mamba \cite{gu2023mamba} further optimized SSM by introducing a selection mechanism and hardware-friendly algorithm design. Mamba selection mechanism enables the model to dynamically adjust parameters based on the input sequence, effectively addressing the discrete modality problem and achieving more efficient processing for long sequences in domains like natural language and genomics. This hardware-optimized and parameter-dynamic strategy not only reduces computational complexity but also improves computational resource efficiency, making Mamba highly effective for long-sequence modeling tasks. Together, the advancements in S4 and Mamba have enhanced the applicability of SSMs in long-sequence data analysis, providing valuable insights for other complex sequence modeling tasks.

\section{Channel-Time Mixing}\label{mixing}
In this section, we first give a brief introduction to the multivariate time series prediction problem and its bottlenecks. We then formulate the problem. Finally, we describe the process of optimally solving the problem.
\subsection{Multivariate Time Series Forecasting}
Multivariate Time Series Forecasting (MTSF) is a method that predicts future values by analyzing time series data consisting of multiple variables. Given a historical multivariate time series input $\mathbf{X} \in \mathbb{R}^{T \times N}$, where $T$ represents the number of time steps and $N$ represents the number of variables, the objective is to predict the future target values for the next $H$ time steps. The prediction sequence is denoted as $\mathbf{\hat{X}} \in \mathbb{R}^{H \times N}$. In MTSF, the data typically contains complex mixed signals across both the temporal and channel dimensions, making it highly challenging to effectively capture meaningful patterns and relationships. The current challenges primarily include:  1) time series signals often contain various mixed components, making feature extraction and unmixing complex; 2) achieving stable mapping between historical and future sequences, as current features cannot be expressed in an explicit mapping across the entire sequence. To address these challenges, we propose a method called MTS-UNMixing, which employs unmixing techniques to mitigate these issues.

\subsection{Problem Formulation}
To solve the problem in MTSF, we formulate the sequence in an equation, that is, we build a mixed model to obtain the main components of each dimension. The components are shared by explicitly mapping them to the sequence, and finally an optimisation equation is proposed for solving.

\paragraph{Mixing Model}
In multivariate time series data, each variable exhibits different feature patterns along the temporal axis. Consider a multivariate time series with \( T \) time steps and \( N \) variables, represented as \( X = \{ x_i \}_{i=1}^T \in \mathbb{R}^N \). 

According to the temporal mixing model, the observation at a given time step consists of several primary temporal patterns (referred to as basis signals). The data at a single time step can be expressed as:

\begin{equation}
	x_i = \sum_{k=1}^{k_1} s_{i k} a_{k} = A_t s_{i, t},
\end{equation}
where \( k_1 \) represents the number of basis signals, \( \{ a_k \}_{k=1}^{k_1} \) denotes \( k_1 \) basis signals, \( \{ s_{i k} \}_{k=1}^{k_1} \) represents the weight of the \( k \)-th basis signal at time step \( i \), \( A_t = [a_{t, 1}, a_{t, 2}, \ldots, a_{t, k_1}] \in \mathbb{R}^{N \times k_1} \) is the basis matrix containing \( k_1 \) distinct basis signals, and \( s_{i, t} = [s_{i, t, 1}, s_{i, t, 2}, \ldots, s_{i, t, k_1}]^T \in \mathbb{R}^{k_1} \) is the weight vector for time step \( i \).

For \( T \) time steps, the matrix form can be defined as:

\begin{equation}
	X = A_t S_t,
\end{equation}
where \( S_t = [s_{1, t}, s_{2, t}, \ldots, s_{T, t}] \in \mathbb{R}^{k_1 \times T} \) is the coefficient matrix. To ensure physical interpretability, the coefficient matrix must satisfy the following constraints:
\begin{itemize}
	\item \textbf{Sum-to-one constraint}: The coefficients for each time step must sum to 1, i.e., 
	\begin{equation}
		E_{k_1}^T S_t = E_T^T,
	\end{equation}
	where \( E_{k_1} = [1, 1, \ldots, 1]^T \) is a column vector with all elements equal to 1, and \( [\cdot]^T \) denotes the transpose operation.
	\item \textbf{Non-negativity constraint}: All elements of the coefficient matrix must be non-negative, i.e., 
	\begin{equation}
		S_t \geq 0.
	\end{equation}
\end{itemize}

Thus, the temporal mixing model can be rewritten as:

\begin{equation}
	X = A_t S_t \quad \text{s.t.} \quad E_{k_1}^T S_t = E_T^T, \quad S_t \geq 0.
\end{equation}

Similar to the time-domain mixture model, the channel mixture model considers the relationship between different channels at each time step. The data of each channel is expressed as a linear combination of several primary channel patterns (base signals), which is expressed as:
\begin{equation}
	X = A_c S_c \quad \text{s.t.} \quad E_{k_2}^T S_c = E_N^T, \quad S_c \geq 0,
\end{equation}
where \( A_c \in \mathbb{R}^{T \times k_2} \) is the basis matrix containing \( k_2 \) primary channel patterns, and \( S_c \in \mathbb{R}^{k_2 \times N} \) is the coefficient matrix, describing the contributions of each basis signal to different channels.
The sum-to-one constraint \( E_{k_2}^T S_c = E_N^T \) ensures the relative contributions of the basis signals, while the non-negativity constraint \( S_c \geq 0 \) guarantees physical interpretability.

\paragraph{Explicit Mapping}

To establish a clear and interpretable relationship between historical and future sequences, we propose an explicit mapping mechanism. This mechanism extracts global features in the temporal and channel dimensions to enable the sharing of key components. Explicit mapping assumes that historical and future sequences can be represented through a unified mixing model to capture the overall patterns. Specifically, the sequences can be expressed as:
\begin{equation}
	X = \begin{bmatrix} A_c', \hat{A_c} \end{bmatrix} S_c, \quad X = A_t \begin{bmatrix} S_t', \hat{S_t} \end{bmatrix},
	\label{mix}
\end{equation}
where \( A_c' \) and \( \hat{A_c} \) are the channel basis matrices for the historical and future sequences, respectively. \( S_t' \) and \( \hat{S_t} \) are the temporal coefficient matrices for the historical and future sequences. \( S_c \) is the shared channel coefficient matrix, ensuring consistent inter-variable relationships between the historical and future sequences. \( \hat{A_c} \in \mathbb{R}^{H \times k_2} \) is the channel basis matrix for the future sequence. Similarly, \( A_t \) is the shared temporal basis matrix, capturing primary temporal features and ensuring consistency in trends and periodicity across the sequences. \( \hat{S_t} \in \mathbb{R}^{k_1 \times H} \) is the temporal coefficient matrix for the future sequence, describing its temporal feature distribution. This unified representation enables consistent modeling of both historical and future sequences and achieves explicit feature decomposition in both dimensions.

Based on the equation \ref{mix}, the historical sequence can be reconstructed as:
\begin{equation}
	X_h = A_t S_t', \quad X_h = A_c' S_c,
		\label{mix2}
\end{equation}
where \( A_t \) is the shared temporal basis matrix, capturing trends and periodic patterns in the historical sequence. \( S_t' \) is the temporal coefficient matrix, describing the contribution of each temporal feature in the historical data. Similarly, \( A_c' \) is the channel basis matrix, reflecting major patterns among channels, and \( S_c \) is the shared channel coefficient matrix, indicating the mixing weights for variables.

The future sequence is predicted as:
\begin{equation}
	\hat{X} = A_t \hat{S_t}, \quad \hat{X} = \hat{A_c} S_c,
		\label{mix3}
\end{equation}
where shared components ensure continuity between historical and future features. The temporal basis \( A_t \) maintains consistency in trends and periodic patterns, while the channel coefficient \( S_c \) ensures stable inter-variable relationships.

The explicit mapping mechanism allows for a clear decoupling of features in the time and channel dimensions. Shared components ensure the continuity of features between historical and future sequences. In the time dimension, the shared basis matrix (\(A_t\)) ensures the consistency of trends and cyclical patterns, thereby enhancing the ability of the model to capture correlations. In the channel dimension, the shared coefficient matrix (\(S_c\)) maintains the stability of the relationship between variables and reduces redundancy during channel mixing. This explicit mapping design improves the interpretability of the model and enhances the accuracy and robustness of the predictions.

\begin{figure*}[ht]
	\centering
	\includegraphics[width=1.0\textwidth, trim=1cm 1.8cm 3.6cm 1.8cm, clip]{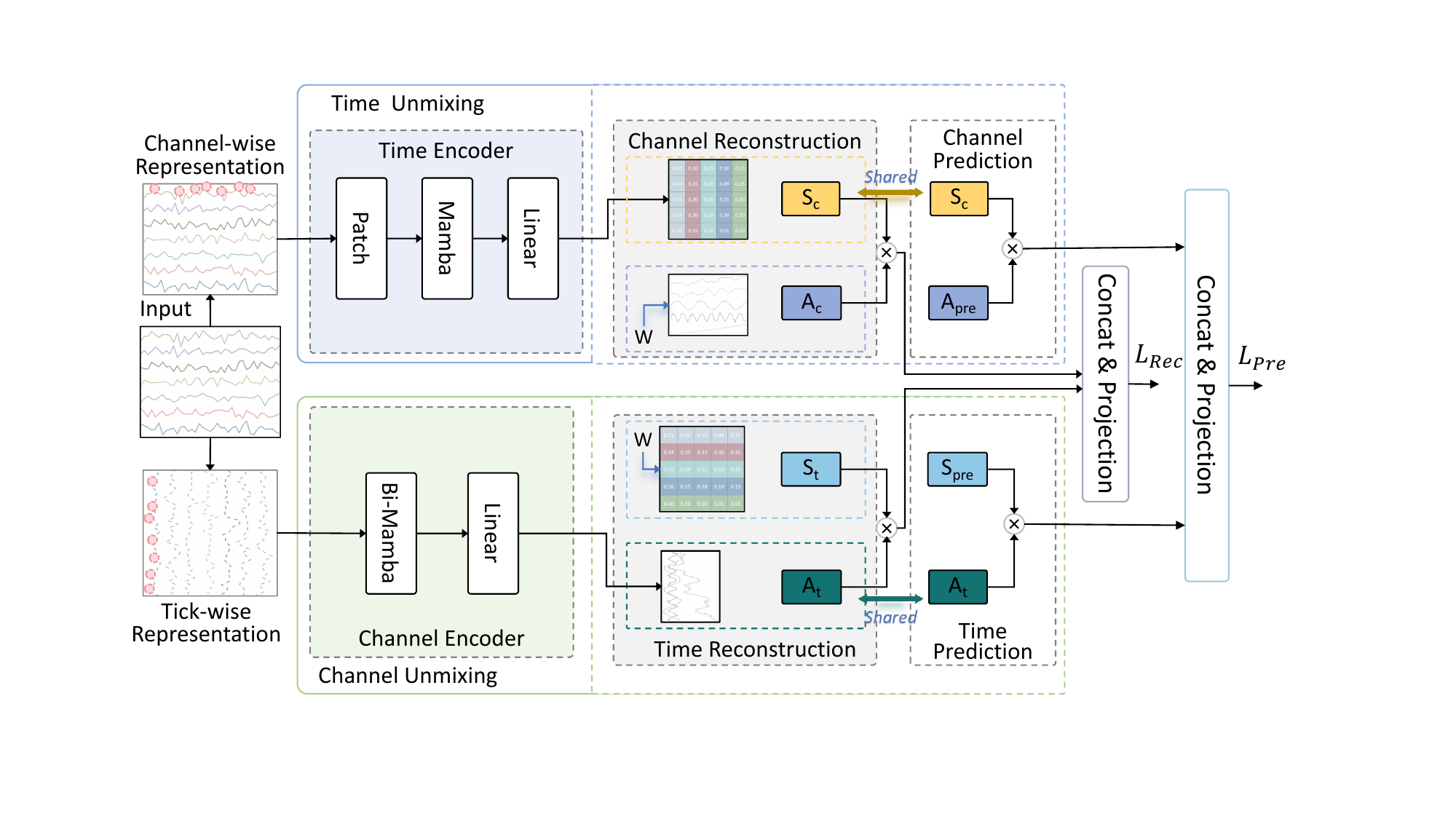}
	\caption{The framework of MTS-UNMixers comprises two main components: temporal unmixing and channel unmixing.}
	\label{method1}
\end{figure*}

\paragraph{Optimization Process} 
To accurately model the feature relationships in both historical and future sequences, we need to estimate both the coefficient matrix and the basis matrix. Specifically, the goal is to minimize the reconstruction error to effectively capture the temporal and channel characteristics of the data. Our objective function is defined as:
\begin{equation}
	(A, S) = \arg\min_{A, S} \| X - AS \|_1,
\end{equation}
where \( \| \cdot \|_1 \) denotes the L1 norm, used here to measure reconstruction error. Since we effectively extract features from both the temporal and channel axes, the optimization process is integrated into the following form:
\begin{equation}
	\begin{aligned}
		(A_c, S_c, A_t, S_t) &= \arg\min_{A_c, S_c} \| X - A_c S_c \|_1 \\
		&\quad + \arg\min_{A_t, S_t} \| X - A_t S_t \|_1 \\
		\text{s.t.} \quad  E_{k_1}^T S_t &= E_T^T, \quad E_{k_2}^T S_c = E_N^T.
	\end{aligned}
	\label{equation_8}
\end{equation}

With this formulation, we simultaneously extract features along the temporal and channel dimensions and address both reconstruction and prediction within a unified framework. Ultimately, our optimization task is divided into two parts: reconstructing the historical sequence and modeling the future sequence. To achieve accurate reconstruction and prediction, we only need to estimate the coefficient matrix and basis matrix for both historical and future sequences, ensuring feature consistency across different time points and variables.

\section{MTS-UNMixers Network}\label{unmxing}
This section provides a detailed overview of the proposed method. We first present the overall architecture, followed by in-depth descriptions of three key components: the unmixing network, the prediction layer, and the loss function.

\subsection{Overall Architecture}
To address the issues outlined in the previous section (equation \ref{equation_8}), we propose a network called MTS-UNMixers, which utilizes Mamba blocks for unmixing. The overall framework is shown in Fig. \ref{method1}. This design includes two main dimensions: the horizontal dimension, which represents each time frame (channel-wise representation) and each channel (tick-wise representation), and the vertical dimension, which represents the stages of model unmixing and reconstruction-prediction.
The first main path focuses on the temporal aspect. In the unmixing stage, Mamba blocks extract complex dynamic patterns and main features over long time steps using nonlinear computation, resulting in a sequence-based coefficient matrix. This matrix is then multiplied with the basis signals to obtain both the reconstructed and predicted sequences.
The second main path addresses the channel aspect. Here, the unmixing stage employs bidirectional Mamba blocks, performing deep calculations in different directions to extract low-dimensional similar features between channels. This results in a shared basis signal for both historical and future sequences. To ensure consistent patterns between different channels in historical and future sequences, the basis signals are multiplied by reconstruction and prediction coefficient matrices, respectively, to produce the final reconstructed and predicted results.
Finally, the outputs from the temporal and channel paths are fused to produce the final output.

\begin{figure}[h]
	\centering
	\rotatebox{-90}{ 
		\includegraphics[width=0.3\textwidth, trim=11cm 2cm 11cm 3cm, clip]{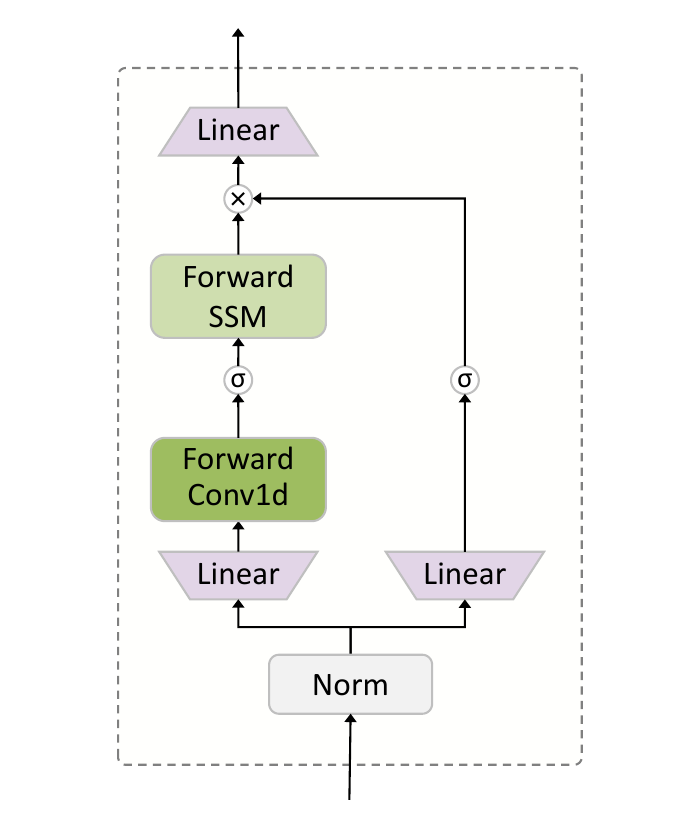}
	}
	\caption{Structure of the Mamba Block.}  
	\label{fig:method2}
\end{figure}

\subsection{Unmixing Encoder} 
\paragraph{Temporal (Channel-wise Representation) Unmixing Encoder} 
The Time Encoder first processes the historical sequence through a blocking operation,  resulting in \( X_{p} \in \mathbb{R}^{N \times T_p \times P} \), where the sequence is divided into small segments (tokens) for feature extraction. Subsequently, these tokens are processed by the Mamba block, which leverages a SSM to capture complex temporal dependencies. As shown in Fig. \ref{fig:method2}. The Mamba block is designed to extract long-range features across time steps, effectively handling non-linear dynamics within the sequence. Additionally, Mamba enhances computational efficiency through parallelized processing, enabling efficient modeling while preserving crucial information. The core equations are as follows:

\begin{equation}
	h'(t) = A h(t) + B x(t), \quad y(t) = C h(t),
\end{equation}
where \( A \) describes the state transition, \( B \) integrates the input signal, and \( C \) maps the state to the output. To handle discrete data, the Zero-Order Hold (ZOH) method is applied for discretization:

\begin{equation}
	h_t = \tilde{A} h_{t-1} + \tilde{B} x_t, \quad y_t = C h_t.
\end{equation}
Mamba introduces the Selective State Space Model (S6), which parameterizes the matrices \( A \), \( B \), and \( C \) dynamically, enabling the model to adapt to complex time series tasks. The final Mamba model is formalized as:

\begin{equation}
	X_m = SSM(Conv(Linear(X_p))) * \sigma(Linear(X_p)),
\end{equation}
where \( \sigma \) is a nonlinear activation function, and \( X_m \) represents the processed output. The encoded output is then passed through a linear layer followed by a softmax operation to generate the weight matrix \( S_c \):

\begin{equation}
	S_c = {Softmax}({Linear}(X_m)).
\end{equation}

\begin{figure}[h]
	\centering
	\rotatebox{-90}{ 
		\includegraphics[width=0.3\textwidth, trim=10cm 2cm 10cm 1cm, clip]{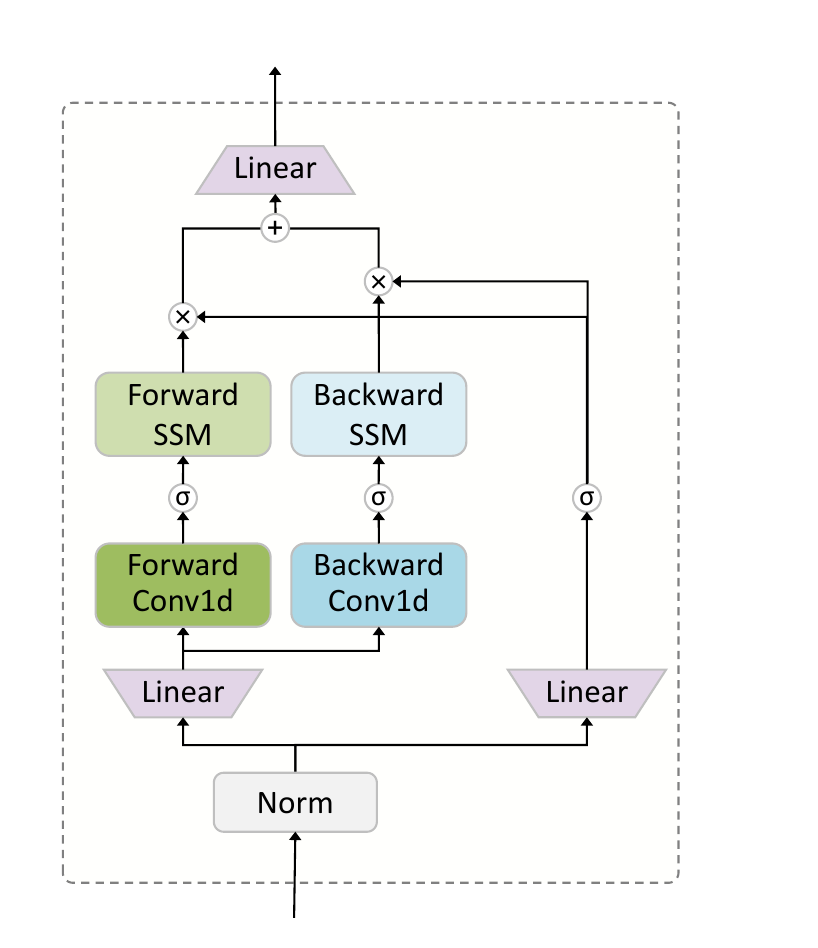}
	}
	\caption{Structure of the Bi-Mamba Block.}  
	\label{fig:method3}
\end{figure}

\paragraph{Channel (Tick-wise Representation) Unmixing Encoder} 
The channel encoder processes the historical sequence across channels at each time step using a bidirectional Mamba block, as shown in Fig. \ref{fig:method3}. The bidirectional Mamba block leverages a bidirectional state space model (SSM) to capture complex dependencies between channels, allowing the model to obtain complete contextual information at each time step. Since there is typically no causal relationship between channels, but rather bidirectional interactions, the bidirectional Mamba is better equipped to comprehensively capture the complex interdependencies among variables. Compared to linear models, the Mamba block enhances precision in capturing complex inter-channel interactions through nonlinear feature extraction. The main operations are:

\begin{equation}
	\begin{aligned}
		X_f &= SSM(Conv(X_{\text{forward}})) * \sigma(X), \\
		X_b &= SSM(Conv(X_{\text{backward}})) * \sigma(X), \\
		X_e &= {Linear}(X_f + X_b),
	\end{aligned}
\end{equation}
where \( X \) represents the linearly transformed historical sequence, and \( X_{\text{forward}} \) and \( X_{\text{backward}} \) denote the forward and reverse inputs of \( X \), respectively. \( X_f \) and \( X_b \) are the forward and backward temporal features processed through the bidirectional Mamba block, while \( X_e \) is the combined encoding output. The function \( \sigma \) is a nonlinear activation function. Here, we use $ReLU$ to introduce nonlinearity and enhance the expressiveness of the model.  Specifically, \( X \) represents the linearly transformed historical sequence.  This encoded result, \( X_e \), is used to generate the temporal basis matrix \( A_t \), which captures the primary temporal patterns of the sequence.

\subsection{Unmixing Decoder}
The unmixing decoder reconstructs the historical sequence and predicts the future sequence using the extracted features, followed by a projection layer to generate the final model output.
\paragraph{Channel Reconstruction and Prediction}
In this phase, the Time Encoder generates the component coefficient matrix \( S_c \), where \( S_c = \text{softmax}(S_c) \), representing the mixing ratio of variables across feature patterns. Based on Equations \ref{mix}, \ref{mix2}, \ref{mix3}, the historical and future sequences are treated as a unified sequence, resulting in a shared \( S_c \).

For reconstruction and prediction, the shared \( S_c \) is used as follows:
\begin{equation}
	X_c' = A_c S_c, \quad \hat{X_c} = A_p S_c,
\end{equation}
where \( A_c \) and \( A_p \) are the learnable basis signal matrices for reconstruction and prediction, respectively. This ensures consistency between historical and future features while enabling accurate representation and forecasting within the channel-wise representation.
\paragraph{Time Reconstruction and Prediction} 
Based on Equations \ref{mix}, \ref{mix2} ,\ref{mix3}, the shared basis signal matrix \( A_t \) is generated through the Channel Encoder to represent the core features of each channel in the temporal dimension. During the unmixing process, the historical and future sequences are treated as a unified sequence, resulting in a consistent basis signal matrix \( A_t \) across both historical and future sequences, thereby forming a shared matrix with a physically meaningful feature pattern representation.

For reconstruction and prediction, the shared \( A_t \) is used as follows:
\begin{equation}
	X_t' = A_t S_t, \quad \hat{X_t} = A_t S_p,
\end{equation}
where \( S_t \) and \( S_p \) are the learnable coefficient matrices for reconstruction and prediction, respectively. Both \( S_t \) and \( S_p \) undergo a \( \text{softmax} \) operation to ensure that their elements represent the relative contributions across different time steps. This approach enables reconstruction and prediction in the temporal dimension while ensuring consistency of temporal features between historical and future sequences, thereby better capturing the primary temporal patterns of the series.

\paragraph{Projection Layer}
In obtaining the prediction results from the temporal and channel axes, the two are concatenated, and a linear layer is applied to generate the final output:

\begin{equation}
	\hat{X} = Linear(Concat(\hat{X_c}, \hat{X_t})).
\end{equation}
 Through concatenation and linear transformation, the model integrates features from both dimensions to enhance prediction accuracy.
Similarly, the reconstructed data undergoes fusion to obtain the reconstructed historical sequence:

\begin{equation}
	{X}^{'} = Linear(Concat({X}_{c}^{'}, {X}_{t}^{'} )).
\end{equation}
The linear layer processes the fused data to generate the final reconstructed sequence.

\subsection{Loss Function}

Based on the optimization formulas \ref{equation_8} presented earlier, we employ the L1 loss function to solve the sequence reconstruction and prediction tasks by measuring the absolute differences between predicted and actual values. Compared to L2 loss, L1 loss is more robust to outliers, making it better suited for capturing the main patterns and trends in time series data.
The overall L1 loss function, incorporating both reconstruction and prediction objectives, is defined as follows:
\begin{equation}
	\mathcal{L} = \lambda_1 \cdot \| {X}^{'} - X_{\text{history}} \|_1 + \lambda_2 \cdot \| \hat{X} - X_{\text{future}} \|_1,
\end{equation}
where \( \lambda_1 \) and \( \lambda_2 \) are weighting factors that balance the importance of the reconstruction and prediction tasks.

\begin{table*}[htbp]
	\centering
	\caption{Multivariate time series forecasting results. The input length T = 96 . The best results are highlighted in bold, and the second-best results are underlined.}
	\scalebox{0.72}{\begin{tabular}{cc|cc|cc|cc|cc|cc|cc|cc|cc|cc|cc|cc}
			\toprule
			\multicolumn{2}{c}{\textbf{Model}} & \multicolumn{2}{c}{\textbf{MTS-UNMixers}} &\multicolumn{2}{c}{\textbf{TimeXer}} & \multicolumn{2}{c}{\textbf{TimeMixer}} & \multicolumn{2}{c}{\textbf{PatchTST}} & \multicolumn{2}{c}{\textbf{TimesNet}} & \multicolumn{2}{c}{\textbf{FITS}} & \multicolumn{2}{c}{\textbf{DLinear}} & \multicolumn{2}{c}{\textbf{FEDformer}} & \multicolumn{2}{c}{\textbf{TiDE}} & \multicolumn{2}{c}{\textbf{Stationary}} & \multicolumn{2}{c}{\textbf{Autoformer}} \\
			\midrule
			\multicolumn{2}{c}{\textbf{Metric}}& \textbf{MSE} & \textbf{MAE} & \textbf{MSE} & \textbf{MAE} & \textbf{MSE} & \textbf{MAE} & \textbf{MSE} & \textbf{MAE} & \textbf{MSE} & \textbf{MAE} & \textbf{MSE} & \textbf{MAE} & \textbf{MSE} & \textbf{MAE} & \textbf{MSE} & \textbf{MAE} & \textbf{MSE} & \textbf{MAE} & \textbf{MSE} & \textbf{MAE} & \textbf{MSE} & \textbf{MAE} \\
			\midrule
			\multirow{5}{*}{\rotatebox{90}{ETTh1}} 
			& 96  & \textbf{0.368} & \textbf{0.388} & 0.377 & 0.397 & \underline{0.375} & 0.400 & 0.393 & 0.408 & 0.384 & 0.402 & 0.701 & 0.558 & 0.386 & 0.432 & {0.376} & 0.419 & 0.427 & 0.450 & 0.513 & 0.491 & 0.449 & 0.459 \\
			& 192 & 0.427 & \textbf{0.419} & \underline{0.425} & 0.426 & 0.429 & \underline{0.421} & 0.445 & 0.434 & 0.436 & 0.429 & 0.718 & 0.570 & 0.437 & 0.459 & \textbf{0.420} & 0.448 & 0.472 & 0.486 & 0.534 & 0.504 & 0.500 & 0.482 \\
			& 336 & \textbf{0.443} & \textbf{0.433} & \underline{0.457} & \underline{0.441} & 0.484 & 0.458 & 0.484 & 0.451 & 0.491 & 0.469 & 0.723 & 0.581 & 0.481 & 0.516 & 0.459 & 0.465 & 0.527 & 0.527 & 0.588 & 0.535 & 0.521 & 0.496 \\
			& 720 & \textbf{0.454} & \textbf{0.429} & \underline{0.464} & \underline{0.463} & 0.498 & {0.482} & 0.480 & 0.471 & 0.521 & 0.500 & 0.712 & 0.595  & 0.519 & {0.452} & 0.506 & 0.507 & 0.644 & 0.605 & 0.643 & 0.616 & 0.514 & 0.512 \\
			\cmidrule(lr){2-24}
			&Avg.& \textbf{0.423} & \textbf{0.417} & 0.431 & 0.432 & 0.447 & \underline{0.440} & 0.451 & 0.441 & 0.458 & 0.450 & 0.714 & 0.576 & 0.456 & 0.465 & 0.440 & 0.460 & 0.518 & 0.517 & 0.570 & 0.537 & 0.496 & 0.487 \\
			\midrule
			
			\multirow{4}{*}{\rotatebox{90}{ETTh2}} 
			& 96  & \textbf{0.273} & \textbf{0.327} & \underline{0.289}& \underline{0.340} & \underline{0.289} & 0.341 & 0.294 & 0.343 & 0.340 & 0.374 & 0.353 & 0.387 & {0.333} & 0.476 & 0.346 & 0.388 & {0.304} & 0.359 & 0.476 & 0.458 & 0.346 & 0.388 \\
			& 192 & \textbf{0.344} & \textbf{0.372} &\underline{0.370} & \underline{0.391} & {0.372} & {0.392} & 0.377 & 0.393 & 0.402 & 0.414 & 0.428 & 0.429 & 0.477 & 0.541 & 0.429 & 0.439 & 0.394 & {0.422} & 0.512 & 0.493 & 0.456 & 0.452 \\
			& 336 & \textbf{0.356} & \textbf{0.390} & 0.422 & {0.434} & \underline{0.386} & \underline{0.414} & 0.381 & 0.409 & 0.452 & 0.452 & 0.454 & 0.455 & 0.594 & {0.657} & 0.496 & 0.487 & 0.385 & 0.421 & 0.552 & 0.551 & 0.482 & 0.486 \\
			& 720 & \textbf{0.405} & \textbf{0.426} & {0.429} & 0.445 & \underline{0.412} & \underline{0.434} & 0.412 & 0.433 & 0.462 & 0.468 & 0.451 & 0.460 & 0.831 & {0.515} & 0.463 & 0.474 & 0.463 & 0.475 & 0.562 & 0.560 & 0.515 & 0.511 \\
			\cmidrule(lr){2-24}
			&AVG. & \textbf{0.345} &\textbf{0.379} & 0.378 & 0.403 & \underline{0.365} & \underline{0.395} & 0.366 & 0.395 & 0.414 & 0.427 & 0.422 & 0.433 & 0.559 & 0.547 & 0.434 & 0.447 & 0.387 & 0.419 & 0.526 & 0.516 & 0.450 & 0.459 \\
			\midrule
			
			\multirow{4}{*}{\rotatebox{90}{ETTm1}} 
			& 96  & \underline{0.317} & \textbf{0.350} & \textbf{0.309} & \underline{0.352} & {0.320} & {0.357} & {0.321} & {0.360} & {0.338} & {0.375} & {0.693} & {0.548} & {0.345} & {0.372} & {0.378} & {0.418} & {0.356} & {0.381} & {0.386} & {0.398} & {0.505} & {0.475} \\
			& 192 & \underline{0.360} & \textbf{0.378} & \textbf{0.355} & \underline{0.378} & {0.361} & {0.381} & {0.362} & {0.384} & {0.374} & {0.387} & {0.710} & {0.557} & {0.380} & {0.389} & {0.426} & {0.441} & {0.391} & {0.399} & {0.459} & {0.444} & {0.553} & {0.496} \\
			& 336 & \textbf{0.388} & \textbf{0.400} & \underline{0.387} & \underline{0.399} & {0.390} & {0.404} & {0.392} & {0.402} & {0.410} & {0.411} & {0.722} & {0.566} & {0.413} & {0.413} & {0.445} & {0.459} & {0.424} & {0.423} & {0.495} & {0.464} & {0.621} & {0.537} \\
			& 720 & \textbf{0.454} & \textbf{0.429} & \underline{0.448} & \underline{0.435} & {0.454} & {0.441} & {0.461} & {0.439} & {0.478} & {0.450} & {0.746} & {0.581} & {0.474} & {0.453} & {0.543} & {0.490} & {0.480} & {0.456} & {0.585} & {0.516} & {0.671} & {0.561} \\
			\cmidrule(lr){2-24}
			& AVG. & \textbf{0.380} & \textbf{0.389} & \textbf{0.375} & \underline{0.391} & {0.381} & {0.396} & {0.384} & {0.396} & {0.400} & {0.406} & {0.718} & {0.563} & {0.403} & {0.407} & {0.448} & {0.452} & {0.413} & {0.415} & {0.481} & {0.456} & {0.588} & {0.517} \\

			\midrule
			\multirow{4}{*}{\rotatebox{90}{ETTm2}} 
			& 96  & \textbf{0.170} & \textbf{0.250} & \underline{0.171} & \underline{0.255} & {0.175} & {0.258} & {0.178} & {0.260} & {0.187} & {0.267} & {0.229} & {0.307} & {0.193} & {0.292} & {0.203} & {0.287} & {0.182} & {0.264} & {0.192} & {0.274} & {0.255} & {0.339} \\
			& 192 & \textbf{0.236} & \textbf{0.293} & {0.238} & {0.300} & \underline{0.237} & \underline{0.299} & {0.249} & {0.307} & {0.249} & {0.309} & {0.284} & {0.337} & {0.284} & {0.362} & {0.269} & {0.328} & {0.256} & {0.323} & {0.280} & {0.339} & {0.281} & {0.340} \\
			& 336 & \textbf{0.297} & \textbf{0.333} & {0.301} & {0.340} & \underline{0.298} &\underline{0.340} & {0.313} & {0.346} & {0.321} & {0.351} & {0.338} & {0.369} & {0.369} & {0.427} & {0.325} & {0.366} & {0.313} & {0.354} & {0.334} & {0.361} & {0.339} & {0.372} \\
			& 720 & \underline{0.392} & \textbf{0.390} & {0.401} & {0.397} & \textbf{0.391} & \underline{0.392} & {0.400} & {0.398} & {0.408} & {0.403} & {0.433} & {0.419} & {0.554} & {0.522} & {0.421} & {0.415} & {0.419} & {0.410} & {0.417} & {0.413} & {0.433} & {0.432} \\
			\cmidrule(lr){2-24}
			& AVG. & \textbf{0.274} & \textbf{0.316} & {0.278} & {0.323} & \underline{0.275} & \underline{0.322} & {0.285} & {0.328} & {0.291} & {0.333} & {0.321} & {0.358} & {0.350} & {0.401} & {0.305} & {0.349} & {0.293} & {0.338} & {0.306} & {0.347} & {0.327} & {0.371} \\

			\midrule
			\multirow{4}{*}{\rotatebox{90}{Weather}} 
			& 96  & \textbf{0.158} & \textbf{0.195} & {0.168} & {0.209} & \underline{0.163} & \underline{0.209} & {0.178} & {0.219} & {0.172} & {0.220} & {0.215} & {0.271} & {0.196} & {0.255} & {0.217} & {0.296} & {0.202} & {0.261} & {0.173} & {0.223} & {0.266} & {0.223} \\
			& 192  & \textbf{0.206} & \textbf{0.242} & {0.220} & {0.254} & \underline{0.208} & \underline{0.250} & {0.224} & {0.259} & {0.219} & {0.261} & {0.264} & {0.305} & {0.237} & {0.296} & {0.276} & {0.336} & {0.242} & {0.298} & {0.245} & {0.285} & {0.307} & {0.285} \\
			& 336  & \underline{0.254} & \textbf{0.286} & {0.276} & {0.294} & \textbf{0.251} & \underline{0.287} & {0.278} & {0.298} & {0.280} & {0.306} & {0.312} & {0.336} & {0.283} & {0.335} & {0.339} & {0.380} & {0.287} & {0.335} & {0.321} & {0.338} & {0.359} & {0.338} \\
			& 720  & \textbf{0.339} & \textbf{0.336} & {0.353} & {0.347} & \textbf{0.339} & \underline{0.341} & {0.353} & {0.346} & {0.365} & {0.359} & {0.381} & {0.377} & {0.345} & {0.381} & {0.403} & {0.428} & {0.351} & {0.386} & {0.414} & {0.410} & {0.419} & {0.410} \\
			\cmidrule(lr){2-24}
			& AVG. & \textbf{0.239} & \textbf{0.265} & {0.254} & {0.276} & \underline{0.240} & \underline{0.272} & {0.258} & {0.281} & {0.259} & {0.287} & {0.293} & {0.322} & {0.265} & {0.317} & {0.309} & {0.360} & {0.271} & {0.320} & {0.288} & {0.314} & {0.338} & {0.314} \\

			\midrule
			\multirow{4}{*}{\rotatebox{90}{Traffic}} 
			& 96 & \underline{0.437} & \textbf{0.274} & \textbf{0.416} & \underline{0.280} & {0.462} & {0.285} & {0.500} & {0.315} & {0.593} & {0.321} & {1.410} & {0.805} & {0.650} & {0.396} & {0.562} & {0.349} & {0.568} & {0.352} & {0.612} & {0.338} & {0.613} & {0.338} \\
			& 192 & \underline{0.454} & \textbf{0.288} & \textbf{0.435} & \textbf{0.288} & {0.473} & {0.296} & {0.498} & {0.299} & {0.617} & {0.336} & {1.413} & {0.806} & {0.598} & {0.370} & {0.562} & {0.346} & {0.612} & {0.371} & {0.613} & {0.340} & {0.616} & {0.340} \\
			& 336 & \underline{0.471} & \textbf{0.295} & \textbf{0.451} & \underline{0.296} & {0.498} & {0.296} & {0.504} & {0.319} & {0.629} & {0.336} & {1.429} & {0.809} & {0.605} & {0.373} & {0.570} & {0.323} & {0.605} & {0.374} & {0.618} & {0.328} & {0.622} & {0.328} \\
			& 720 & \underline{0.504} & \textbf{0.312} & \textbf{0.484} & {0.314} & {0.506} & \underline{0.313} & {0.542} & {0.335} & {0.640} & {0.350} & {1.502} & {0.820} & {0.645} & {0.394} & {0.596} & {0.368} & {0.647} & {0.410} & {0.653} & {0.355} & {0.660} & {0.355} \\
			\cmidrule(lr){2-24}
			& AVG. & \underline{0.466} & \textbf{0.292} & \textbf{0.447} & {0.295} & {0.485} & {0.298} & {0.511} & {0.317} & {0.620} & {0.336} & {1.439} & {0.810} & {0.625} & {0.383} & {0.573} & {0.347} & {0.608} & {0.377} & {0.624} & {0.340} & {0.628} & {0.340} \\
			
			\midrule
			\multirow{4}{*}{\rotatebox{90}{Electricity}} 
			& 96 & \textbf{0.147} & \textbf{0.246} & \underline{0.151} & \underline{0.247} & {0.153} & {0.247} & {0.180} & {0.259} & {0.168} & {0.272} & {0.846} & {0.762} & {0.197} & {0.282} & {0.183} & {0.297} & {0.194} & {0.277} & {0.169} & {0.273} & {0.201} & {0.317} \\
			& 192 & \textbf{0.165} & \underline{0.264} & \textbf{0.165} & \textbf{0.261} & {0.166} & {0.256} & {0.188} & {0.268} & {0.184} & {0.289} & {0.849} & {0.761} & {0.196} & {0.285} & {0.195} & {0.308} & {0.193} & {0.280} & {0.182} & {0.286} & {0.222} & {0.334} \\
			& 336 & \textbf{0.181} & \textbf{0.277} & \underline{0.183} & \underline{0.280} & {0.185} & {0.277} & {0.203} & {0.288} & {0.198} & {0.300} & {0.861} & {0.765} & {0.209} & {0.301} & {0.212} & {0.313} & {0.206} & {0.296} & {0.200} & {0.304} & {0.231} & {0.338} \\	
			& 720 & \textbf{0.210} & \textbf{0.304} & \underline{0.220} & \underline{0.309} & {0.225} & {0.310} & {0.239} & {0.321} & {0.220} & {0.320} & {0.892} & {0.775} & {0.245} & {0.333} & {0.231} & {0.343} & {0.242} & {0.328} & {0.222} & {0.321} & {0.254} & {0.361} \\
			\cmidrule(lr){2-24}			
			& AVG. & \textbf{0.176} & \textbf{0.273} & \underline{0.180} & {0.274} & 0.182 & 0.273 & \underline{0.203} & 0.284 & {0.193} & 0.295 & 0.862 & 0.766 & 0.212 & 0.300 & 0.205 & 0.315 & 0.209 & 0.295 & 0.193 & 0.296 & 0.227 & 0.338 \\
			\midrule
			\multicolumn{2}{c}{\textbf{$1^{\text{st}}$ Count}}&23 & 33 & 9 & 3 & 3 & 0 & 0 & 0 & 0 & 0 & 0 & 0 & 0 & 0 & 1 & 0 & 0 & 0 & 0 & 0 & 0 & 0\\
			
			\bottomrule
	\end{tabular}}
\label{example}
\end{table*}

\section{Experiment Resluts}\label{reslut}
\subsection{Datasets Descriptions}
We conducted experiments on seven real-world datasets to evaluate the performance of the proposed MTS-UNMixers model. These datasets include: (1) ETT dataset (ETTh1, ETTh2, ETTm1, ETTm2), which contains data from two power transformers (from two distinct sites) recorded from July 2016 to July 2018, including variables such as load and oil temperature, totaling seven features. Specifically, ETTm1 and ETTm2 are datasets sampled at a minute-level frequency, while ETTh1 and ETTh2 are sampled at an hourly frequency. (2) Weather dataset, collected by the Max Planck Institute for Biogeochemistry in 2020, which contains 21 meteorological factors, including temperature and humidity, sampled every 10 minutes. This dataset is suitable for evaluating the ability of a model to process complex, multivariate meteorological data. (3) Traffic dataset, which describes the road occupancy rates recorded by 862 sensors on highways in the San Francisco Bay Area from January 2015 to December 2016, with data sampled at an hourly frequency. (4) Electricity dataset, which includes the hourly electricity consumption records of 321 customers from 2012 to 2014. It captures the power usage patterns of different customers over time, serving as a common benchmark dataset for evaluating time series models with high-dimensional multivariate inputs. We followed the same data processing protocol and prediction horizon settings as used in TimesNet\cite{wu2023timesnet}, varying the forecasting length among $\{96, 192, 336, 720\}$ to assess model performance across different forecast windows.
\subsection{Forecasting Performance}
\paragraph{Comparison Methods}
We compared the proposed MTS-UNMixers with nine well-established and advanced models for time series forecasting, including Transformer-based approaches: PatchTST\cite{nietime}, FEDformer\cite{zhou2022fedformer}, Autoformer\cite{wu2021autoformer}, Stationary Transformer\cite{liu2022non}, and TimeXer\cite{wang2024timexer}; MLP-based models: DLinear\cite{zeng2023transformers}, FITS\cite{xufits}, and TiDE\cite{daslong}; as well as CNN-based models: TimesNet\cite{wu2023timesnet} and TimeMixer\cite{TimeMachine}. These models were selected to provide a comprehensive benchmark for evaluating the performance of MTS-UNMixers.

\begin{figure*}[hb]
	\centering
	\captionsetup[subfigure]{labelformat=empty}
	\begin{minipage}{0.22\linewidth}
		\vspace{0.1pt}
		\subcaption{Electricity}{\includegraphics[width=\textwidth]{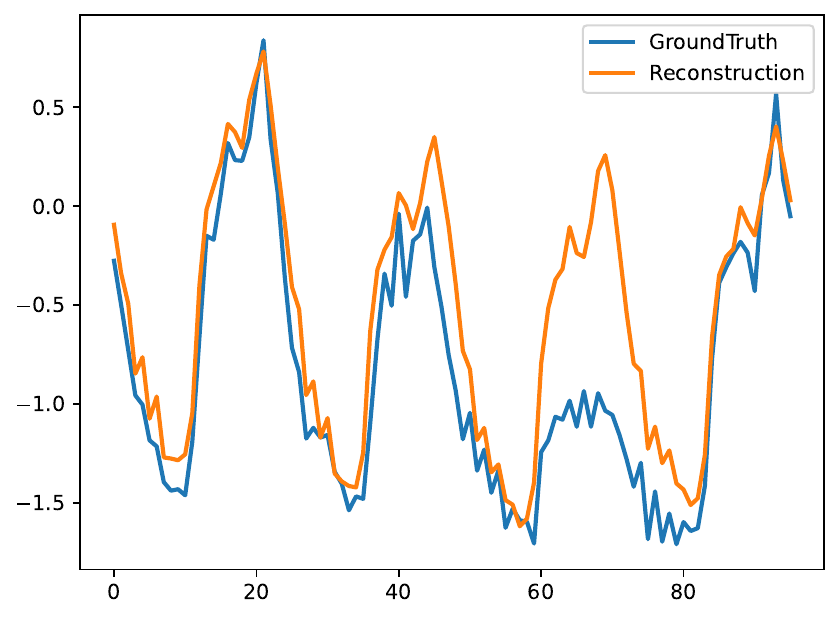}}
		\vspace{0.5pt}
		\subcaption{ETTh1}{\includegraphics[width=\textwidth]{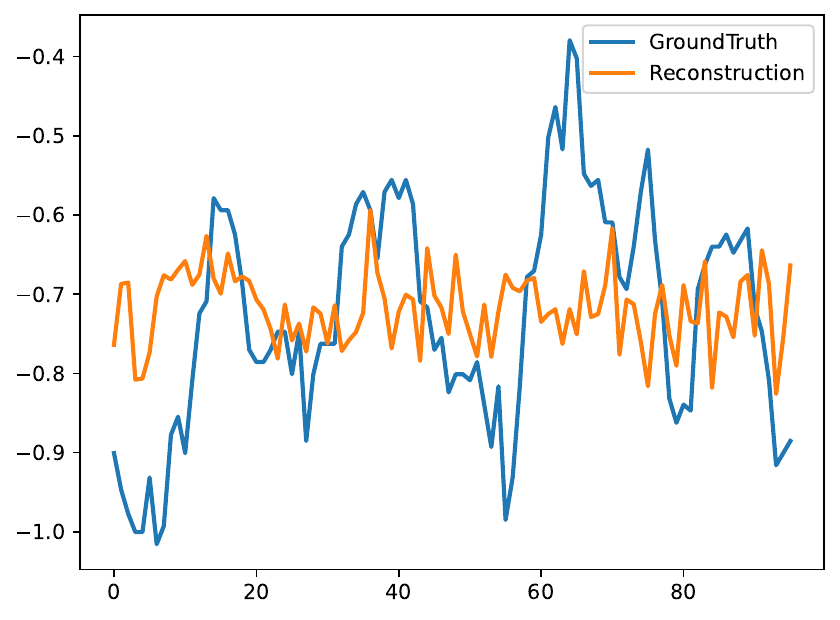}}
		\vspace{0.5pt}
		\subcaption{Weather}{\includegraphics[width=\textwidth]{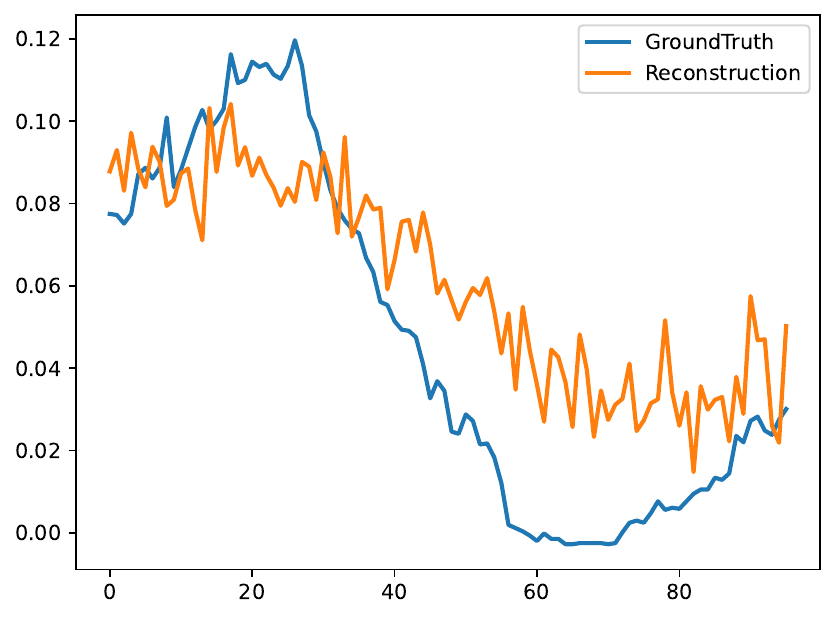}}
		\vspace{0.5pt}
		\centerline{(a) Channel Unmixing}
	\end{minipage}
	\hspace{0.2cm}
	\begin{minipage}{0.22\linewidth}
		\vspace{0.5pt}
		\subcaption{Electricity}{\includegraphics[width=\textwidth]{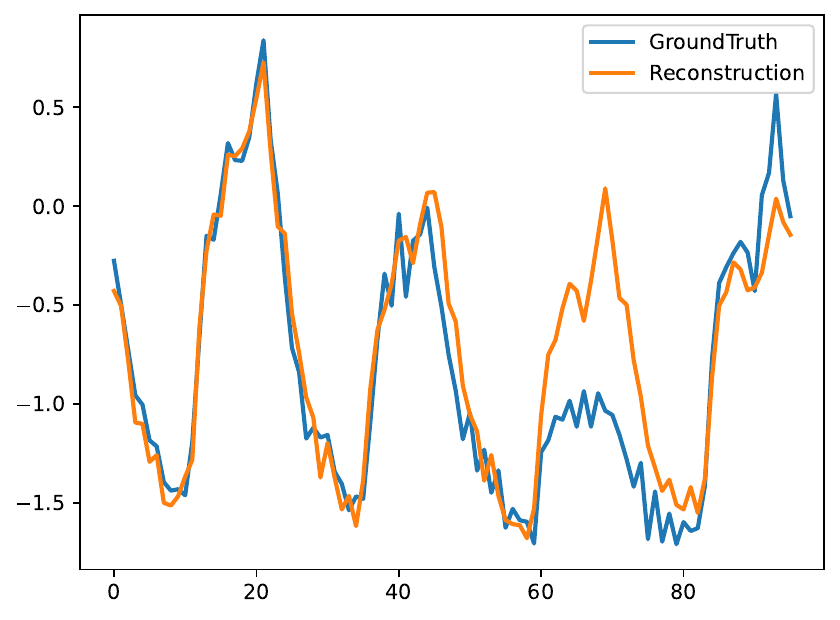}}
		\vspace{0.5pt}
		\subcaption{ETTh1}{\includegraphics[width=\textwidth]{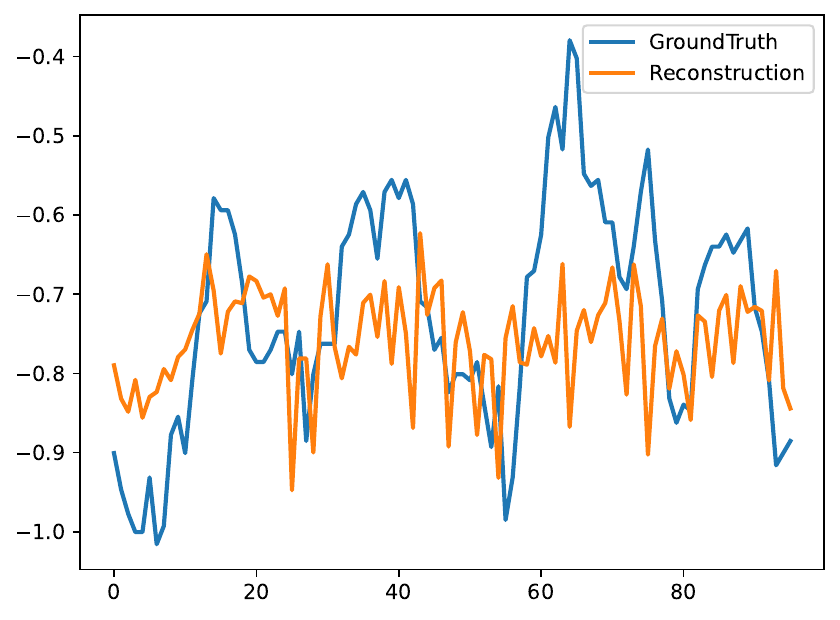}}
		\vspace{0.5pt}
		\subcaption{Weather}{\includegraphics[width=\textwidth]{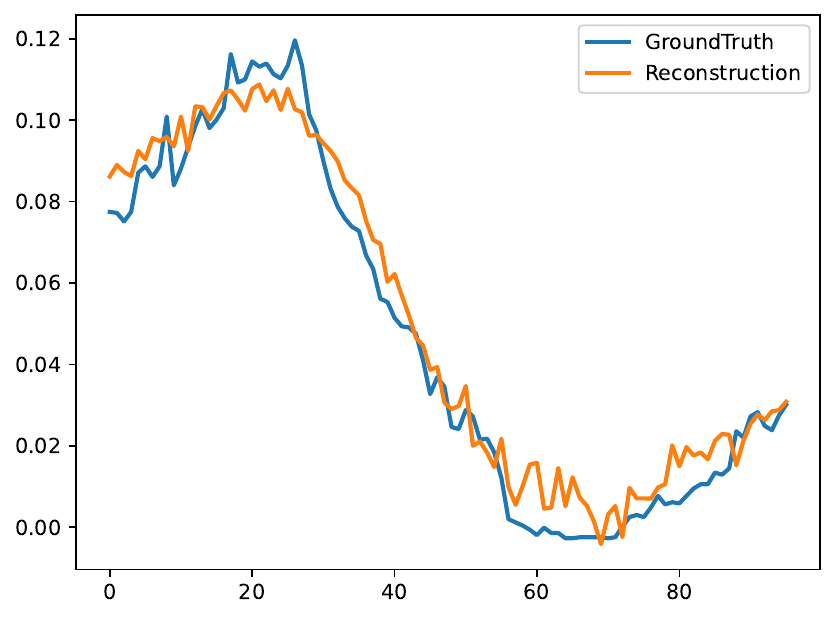}}
		\vspace{0.5pt}
		\centerline{(b) Temporal Unmixing}
	\end{minipage}
	\hspace{0.2cm}
	\begin{minipage}{0.22\linewidth}
		\vspace{0.5pt}
		\subcaption{Electricity}{\includegraphics[width=\textwidth]{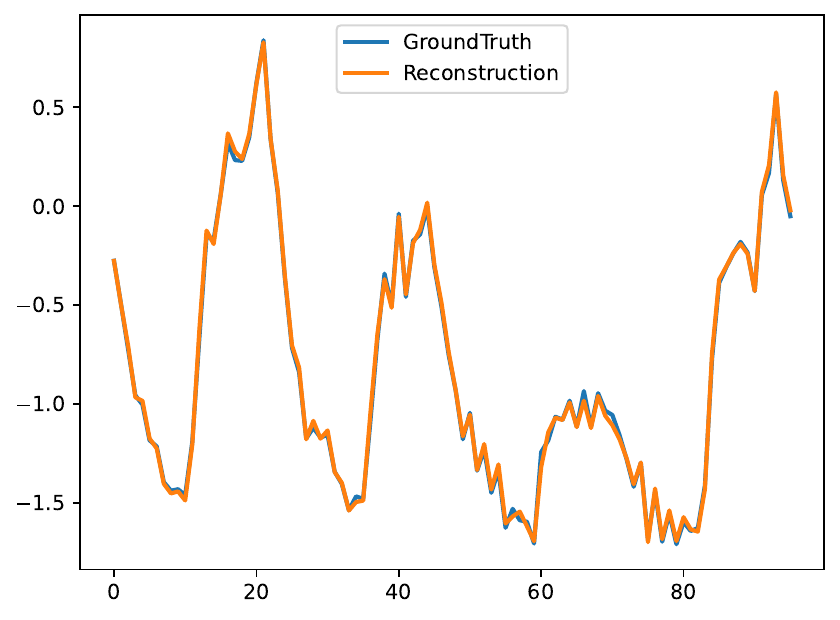}}
		\vspace{0.5pt}
		\subcaption{ETTh1}{\includegraphics[width=\textwidth]{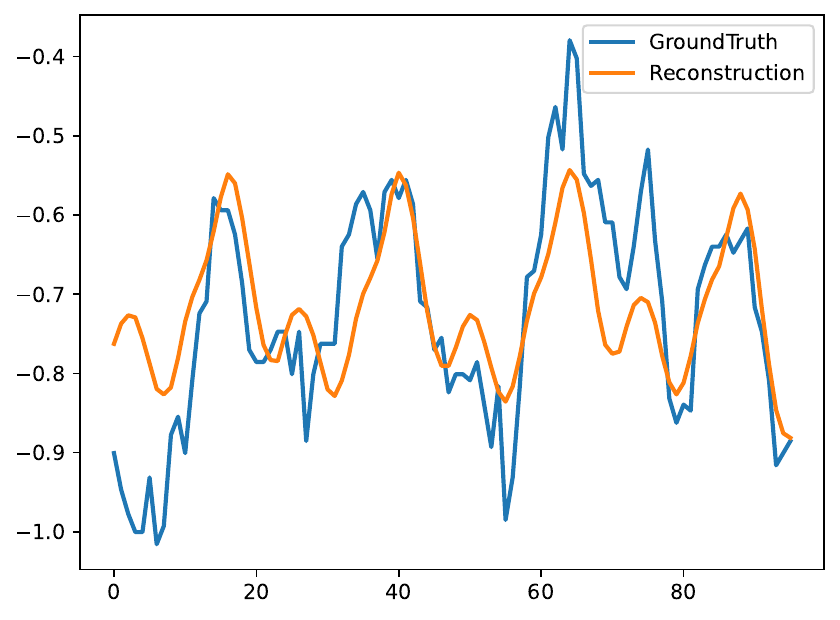}}
		\vspace{0.5pt}
		\subcaption{Weather}{\includegraphics[width=\textwidth]{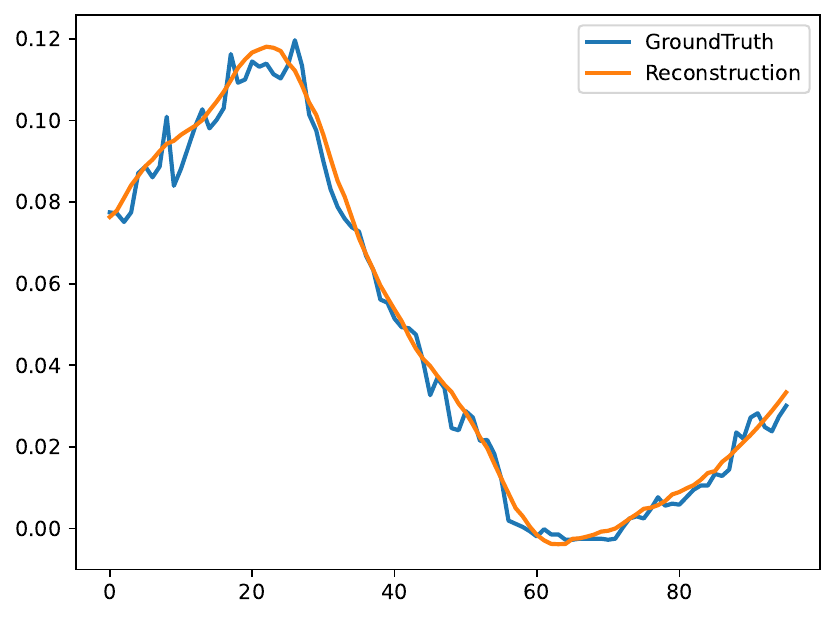}}
		\vspace{0.5pt}
		\centerline{(c) Dual Unmixing}
	\end{minipage}
	\caption{Reconstruction visualization of MTS-UNMixers under different configurations on ETTh1, Weather, and Electricity datasets using a history sequence length of 96. (a) Channel Unmixing Only, (b) Temporal Unmixing Only, (c) Dual Unmixing. Dual-path unmixing achieves the best reconstruction accuracy by effectively capturing both temporal dependencies and inter-channel relationships.}
	\label{abl_unmixing}
\end{figure*}
\paragraph{Main Results}
Table ~\ref{example} summarizes the performance comparison between MTS-UNMixers and the benchmark models across seven datasets. The results show that MTS-UNMixers consistently ranks within the top two in all scenarios, achieving or approaching state-of-the-art performance. Notably, MTS-UNMixers ranked first in 56 cases. Compared to the second-ranking model, TimeMixer, MTS-UNMixers achieved a 5.23\% relative reduction in average MSE and a 5.49\% reduction in MAE on the ETTh1 dataset, as well as a 5.21\% reduction in average MSE on the ETTm2 dataset. These results demonstrate that MTS-UNMixers performs exceptionally well on most datasets, such as ETTh1, ETTm2, and Weather, effectively removing redundant information and reducing sequence complexity. Its strong feature separation capability enables outstanding performance in tasks involving complex temporal dependencies and multivariate interactions. However, on the Traffic dataset, MTS-UNMixers slightly underperforms compared to TimeXer. This may be due to the fact that the traffic dataset has a large number of channels and significant spatial features, while TimeXer excels with its cross-variable attention mechanism and effective use of exogenous variables, which allows it to better capture spatial dependencies and complex variable interactions.
\begin{figure*}[ht]
	\centering
	\captionsetup[subfigure]{labelformat=empty}
	\begin{minipage}{0.23\linewidth}
		\vspace{1pt}
		\subcaption{MTS-UNMixers (Ours)}{\includegraphics[width=\textwidth]{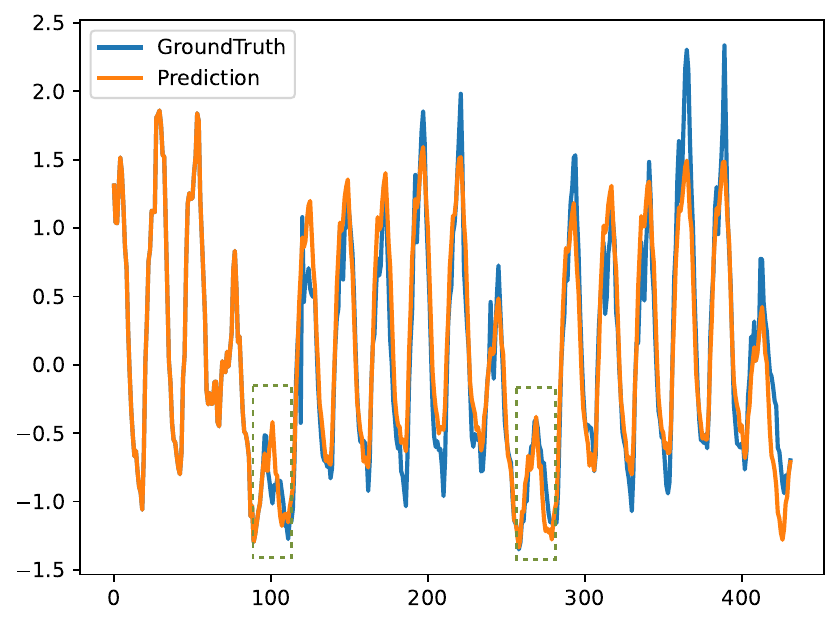}}
		\vspace{1pt}
		\subcaption{Infomer}{\includegraphics[width=\textwidth]{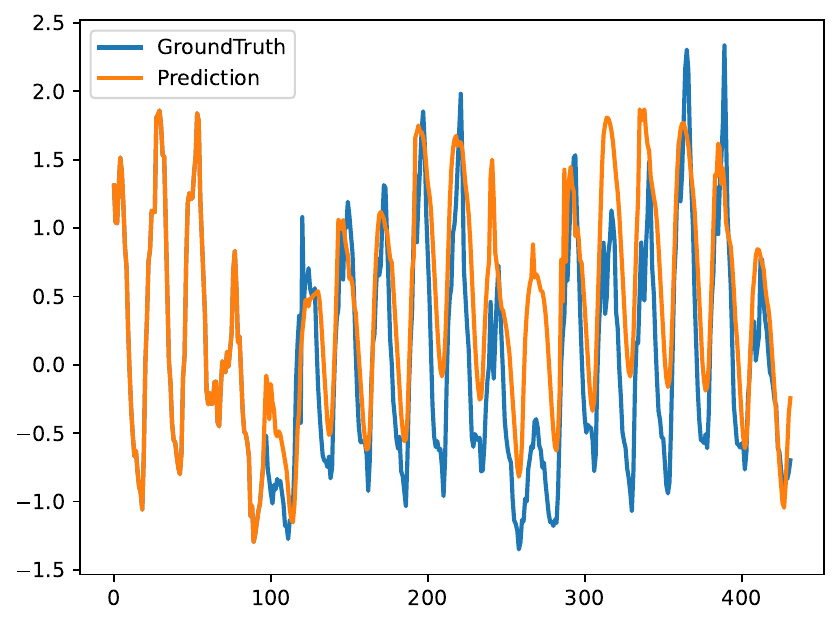}}
		\vspace{1pt}
		\subcaption{MTS-UNMixers (Ours)}{\includegraphics[width=\textwidth]{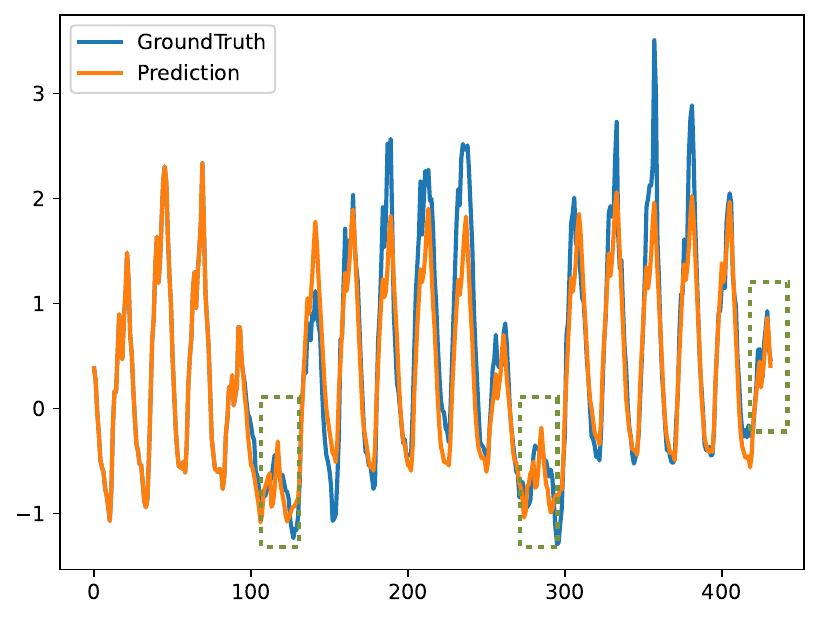}}
		\vspace{1pt}
		\subcaption{Infomer}{\includegraphics[width=\textwidth]{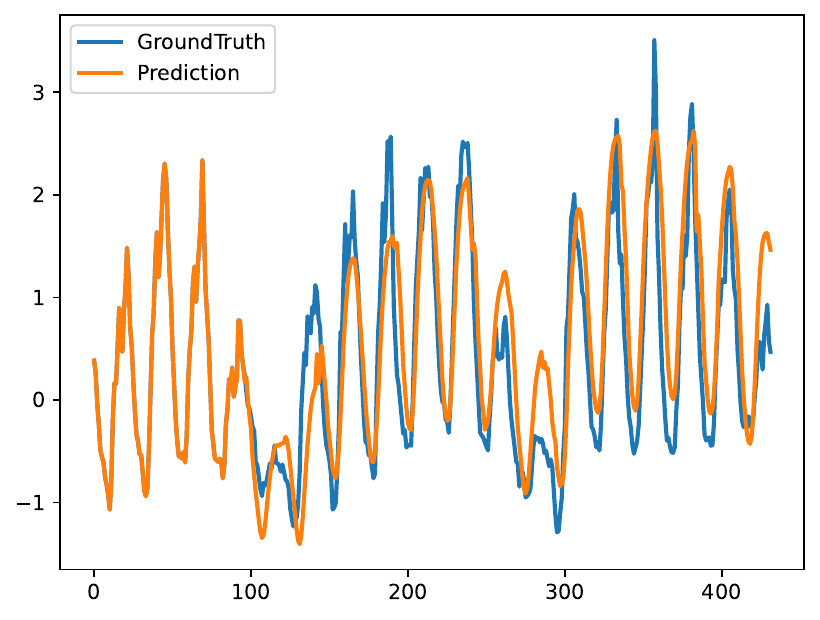}}
	\end{minipage}
	\begin{minipage}{0.23\linewidth}
		\vspace{1pt}
		\subcaption{DLinear}{\includegraphics[width=\textwidth]{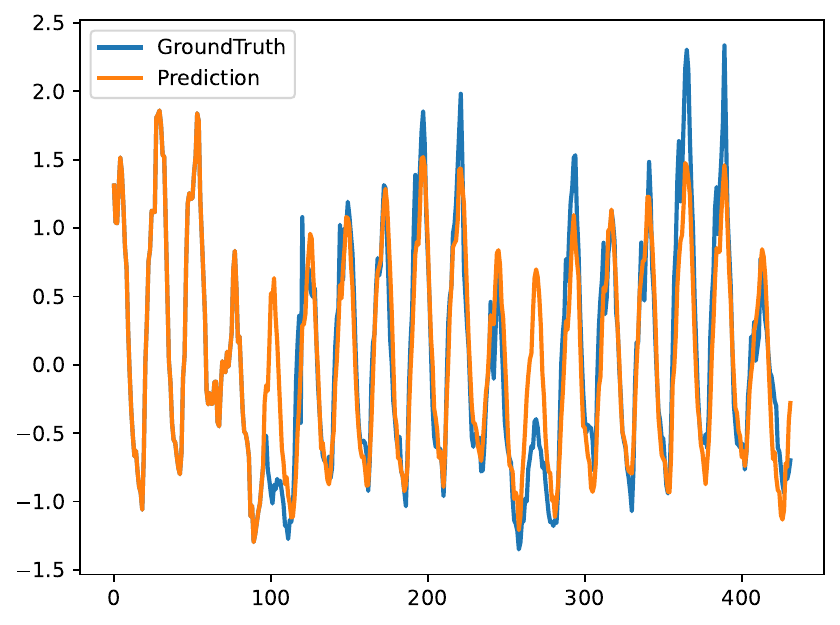}}
		\vspace{1pt}
		\subcaption{Transfomer}{\includegraphics[width=\textwidth]{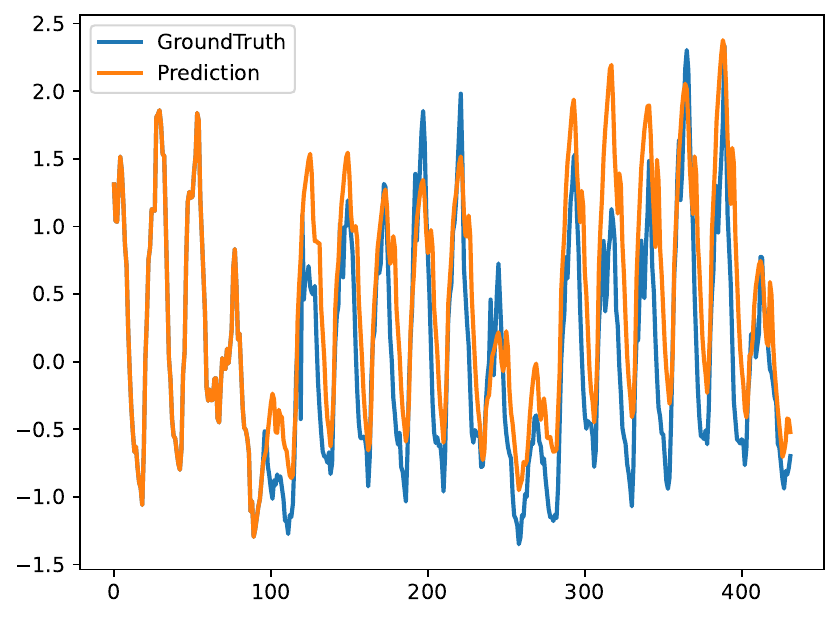}}
		\vspace{1pt}
		\centerline{\quad\quad\quad\quad\quad\quad\quad\quad\quad\quad\quad\quad(a) The First Group.}
		\subcaption{DLinear}{\includegraphics[width=\textwidth]{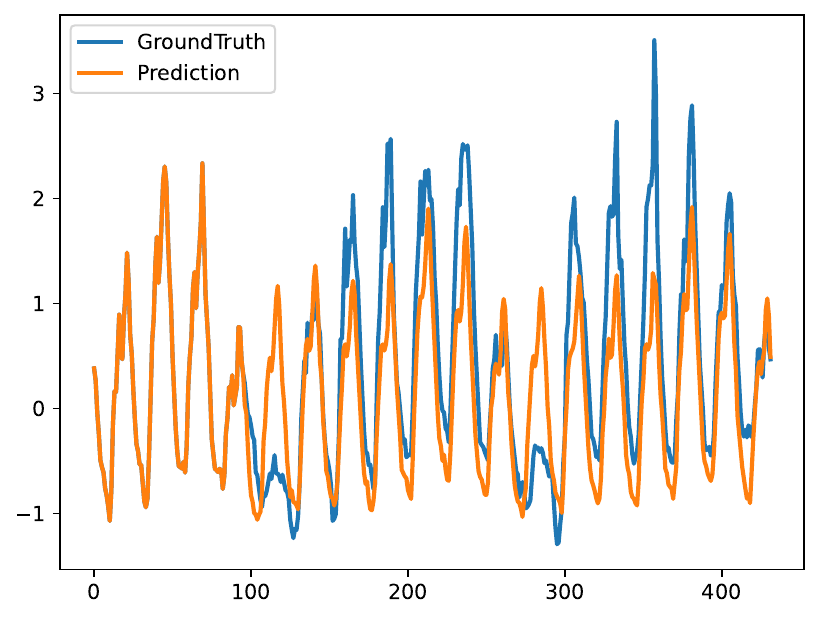}}
		\vspace{1pt}
		\subcaption{Transfomer}{\includegraphics[width=\textwidth]{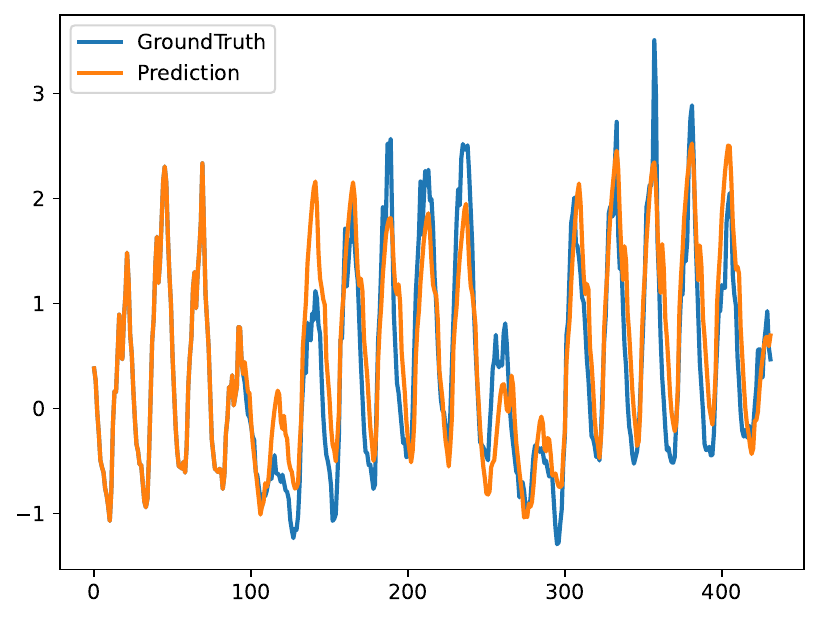}}
	\end{minipage}
	\begin{minipage}{0.23\linewidth}
		\vspace{1pt}
		\subcaption{PatchTST}{\includegraphics[width=\textwidth]{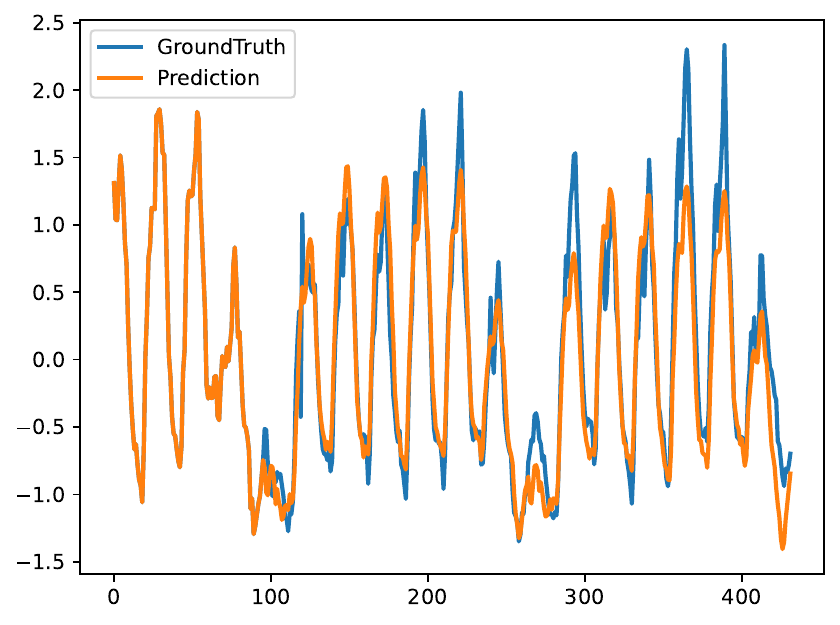}}
		\vspace{1pt}
		\subcaption{TimeXer}{\includegraphics[width=\textwidth]{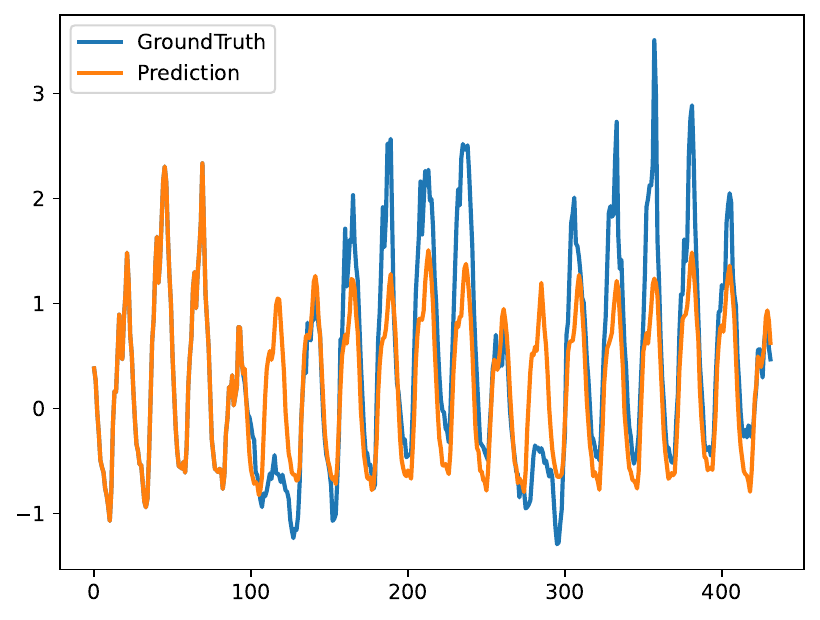}}
		\vspace{1pt}
		\subcaption{PatchTST}{\includegraphics[width=\textwidth]{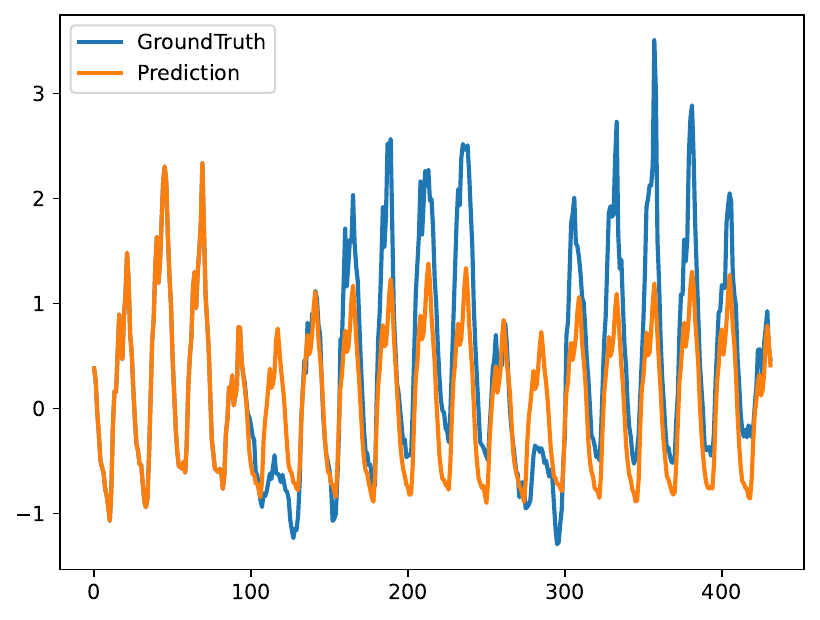}}
		\vspace{1pt}
		\subcaption{TimeXer}{\includegraphics[width=\textwidth]{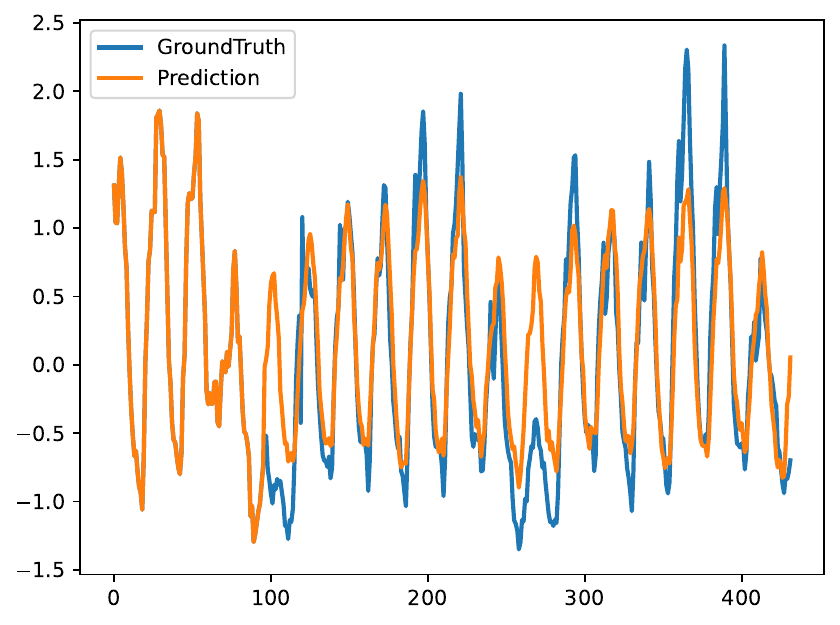}}
	\end{minipage}
	\begin{minipage}{0.23\linewidth}
		\vspace{1pt}
		\subcaption{FEDformer}{\includegraphics[width=\textwidth]{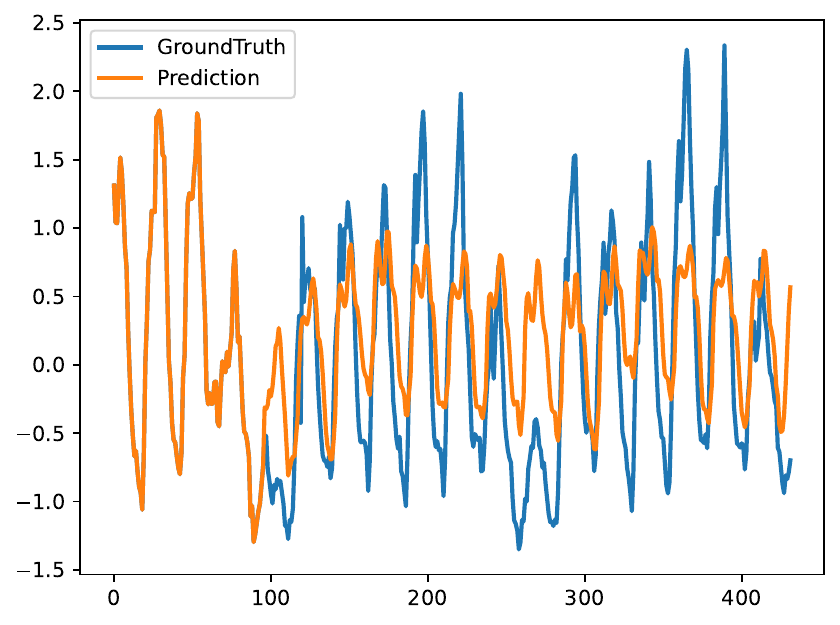}}
		\vspace{1pt}
		\subcaption{TimeNet}{\includegraphics[width=\textwidth]{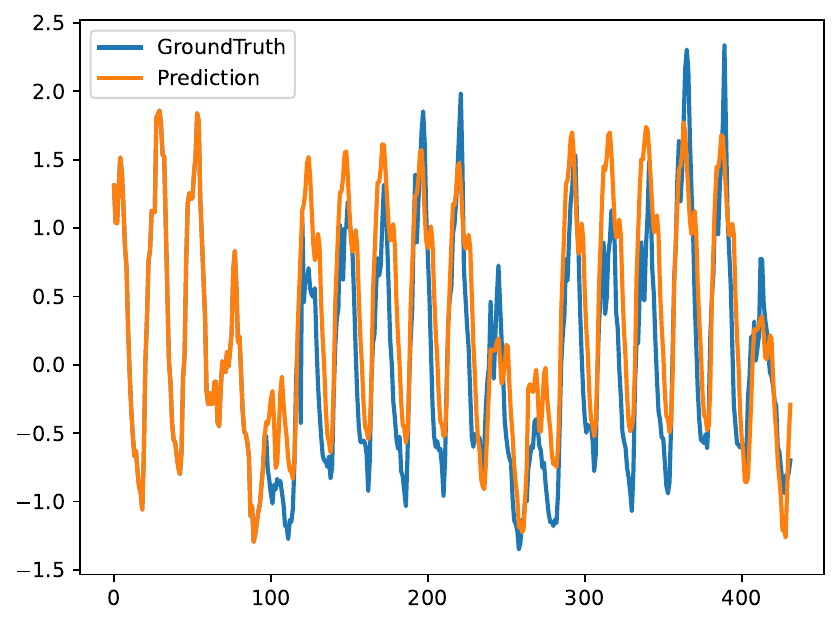}}
		\vspace{1pt}
		\subcaption{FEDformer}{\includegraphics[width=\textwidth]{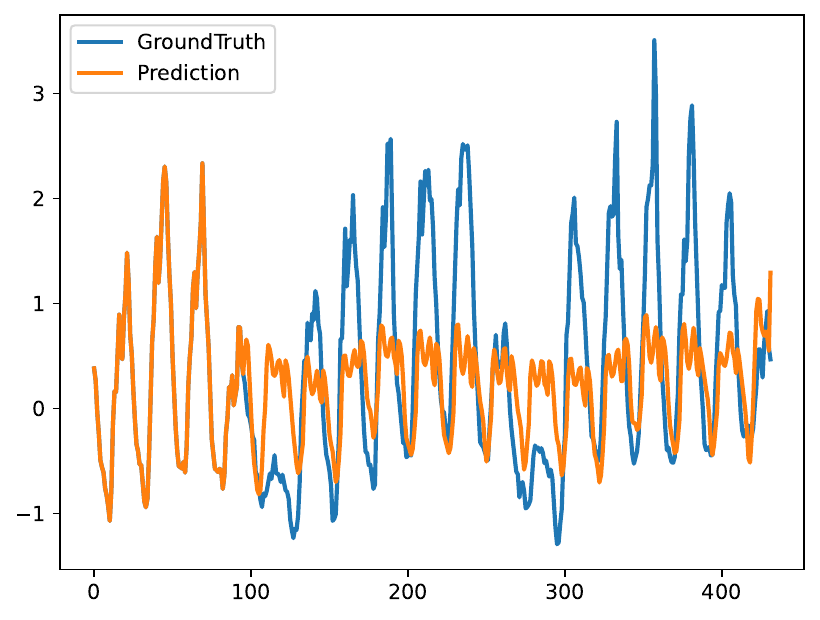}}
		\vspace{1pt}
		\subcaption{TimeNet}{\includegraphics[width=\textwidth]{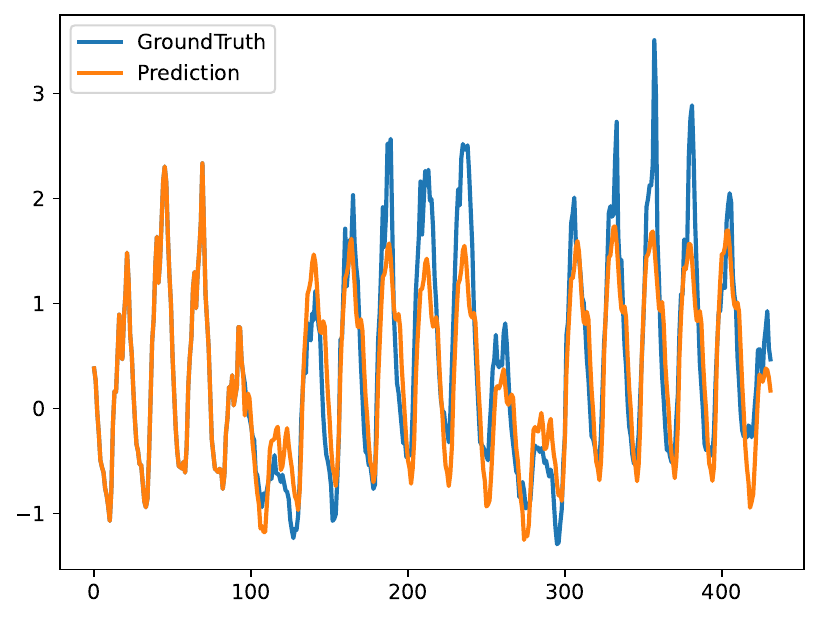}}
	\end{minipage}
	\vspace{4pt} 
	\centerline{(b) The Second Group.}
	
	\caption{Visualization of MTS-UNMixers' prediction results on Electricity. The input length $T = 96$. The results of two experimental groups are provided.}
	\label{Visualization1}
\end{figure*}

\paragraph{Reconstruction visualization}
To better understand the performance of MTS-UNMixers, we designed reconstruction experiments with different module configurations and conducted prediction visualization on the electricity dataset. These experiments utilized a history sequence length of 96 for the ETTh1, Weather, and Electricity datasets, with the following configurations:
\begin{itemize}
	\item Channel Unmixing Only: In this setup, we retained only the channel unmixing module to reconstruct the input data. As shown in Fig. \ref{abl_unmixing} (a), channel unmixing effectively captures inter-channel relationships, allowing feature extraction along the channel dimension. However, without temporal unmixing, the model struggles with time dependencies, leading to visible discrepancies in trend continuity, especially in datasets with long-term dependencies.
	\item Temporal Unmixing Only: In this setup, only the temporal unmixing module was used. Fig. \ref{abl_unmixing} (b) shows that temporal unmixing better captures time dependencies and dynamic trends, resulting in more accurate temporal reconstructions. However, it lacks inter-channel feature extraction, leading to weaker reconstruction of inter-variable interactions.
	\item Dual Unmixing: In this complete configuration, both temporal and channel unmixing modules were applied. As illustrated in Fig. \ref{abl_unmixing} (c), dual unmixing enables the model to deeply extract features across both dimensions, capturing both temporal dependencies and inter-channel details with high accuracy. This dual-path setup demonstrates superior reconstruction compared to single-path configurations.
\end{itemize}

Through a comparative analysis of these visualizations, we observe that dual-path unmixing demonstrates a clear advantage in information extraction and reconstruction. In contrast, the channel-only unmixing configuration performs well in capturing channel-specific features but lacks temporal dependency representation. Meanwhile, the temporal-only configuration fails to capture intricate inter-channel details. By integrating both temporal and channel information, dual-path unmixing achieves optimal accuracy and information integrity in reconstruction, further validating the effectiveness of dual-path unmixing in complex time-series forecasting tasks and providing strong support for the design of MTS-UNMixers.

\paragraph{Prejection visualization}

To provide an intuitive understanding of the forecasting process, we present the prediction results from the electricity dataset in Fig. \ref{Visualization1}. The results of five models are recorded for a 96-input, 336-prediction setting. From the figure, we can observe that our model performs relatively well during cyclical fluctuations. As shown in the Fig. \ref{Visualization1}, MTS-UNMixers can respond accurately to fluctuations in signal generation cycles and trends, with precise predictions, especially in the areas highlighted by the green boxes.

\begin{table}[hb]
	\centering
	\caption{Performance comparison at various prediction lengths of MTS-UNMixers and its ablated versions on the ETTh1, ETTm1, and Weather datasets.}
	\scalebox{0.85}{
		\begin{tabular}{c|c|cc|cc|cc}
			\toprule
			\multirow{2}{*}{Model} & \multirow{2}{*}{Length} & \multicolumn{2}{c|}{ETTh1} & \multicolumn{2}{c|}{ETTm1} & \multicolumn{2}{c}{Weather} \\
			& & MSE & MAE & MSE & MAE & MSE & MAE \\
			\midrule
			\multirow{4}{*}{\textbf{MTS-UNMixers}} & 96 & \textbf{0.368} & \textbf{0.388} & \textbf{0.317} & \textbf{0.350} & \textbf{0.158} & \textbf{0.195} \\
			& 192 & \textbf{0.427} & \textbf{0.419} & \textbf{0.360} & \textbf{0.378} & \textbf{0.206} & \textbf{0.242} \\
			& 336 & \textbf{0.443} & \textbf{0.433} & \textbf{0.388} & \textbf{0.400} & \textbf{0.254} & \textbf{0.286} \\
			& 720 & \textbf{0.454} & \textbf{0.429} & \textbf{0.454} & \textbf{0.429} & \textbf{0.339} & \textbf{0.336} \\
			\midrule
			\multirow{4}{*}{w/o Channel Unmixing} & 96 & 0.375 & 0.393 & 0.321 & 0.355 & 0.159 & 0.196 \\
			& 192 & 0.433 & 0.425 & 0.366 & 0.384 & 0.220 & 0.252 \\
			& 336 & 0.454 & 0.442 & 0.396 & 0.406 & 0.264 & 0.295 \\
			& 720 & 0.466 & 0.436 & 0.461 & 0.432 & 0.346 & 0.350 \\
			\midrule
			\multirow{4}{*}{w/o Time Unmixing} & 96 & 0.477 & 0.458 & 0.383 & 0.364 & 0.204 & 0.270 \\
			& 192 & 0.518 & 0.479 & 0.478 & 0.418 & 0.261 & 0.284 \\
			& 336 & 0.537 & 0.498 & 0.476 & 0.441 & 0.379 & 0.372 \\
			& 720 & 0.547 & 0.516 & 0.560 & 0.530 & 0.471 & 0.372 \\
			\midrule
			\multirow{4}{*}{w/o Mamba} & 96 & 0.369 & 0.393 & 0.320 & 0.351 & 0.162 & 0.226 \\
			& 192 & 0.439 & 0.430 & 0.375 & 0.389 & 0.218 & 0.260 \\
			& 336 & 0.454 & 0.443 & 0.411 & 0.418 & 0.275 & 0.294 \\
			& 720 & 0.461 & 0.448 & 0.465 & 0.449 & 0.352 & 0.356 \\
			\midrule
			\multirow{4}{*}{w/o Bi-Mamba} & 96 & 0.374 & 0.391 & 0.319 & 0.353 & 0.163 & 0.200 \\
			& 192 & 0.433 & 0.428 & 0.366 & 0.381 & 0.213 & 0.248 \\
			& 336 & 0.449 & 0.438 & 0.396 & 0.407 & 0.260 & 0.287 \\
			& 720 & 0.460 & 0.437 & 0.460 & 0.435 & 0.342 & 0.345 \\
			\bottomrule
	\end{tabular}}
	\label{aba1}
\end{table}

\subsection{Ablation Study}
In the ablation study, we systematically evaluated the importance of the main modules within MTS-UNMixers in time-series forecasting, as shown in Table \ref{aba1}. We successively removed the Channel Unmixing, Time Unmixing, Mamba block and Bi-Mamba modules, while keeping other components unchanged. Experiments were conducted on the ETTh1, ETTm1, and Weather datasets, with MSE and MAE recorded at different prediction horizons (96, 192, 336, 720) to assess each module's contribution. 

The results indicated that removing these modules led to significant performance declines, especially for long-term predictions. For example, on the ETTh1 dataset, the MSE for the 96-step forecast dropped by approximately 1.9\% after removing the Channel Unmixing module. When the Time Unmixing module was removed, the MSE for the same horizon dropped significantly by about 29.6\%, highlighting the critical role of Time Unmixing in reducing aliasing errors and capturing temporal dependencies. Overall, the full MTS-UNMixers model consistently achieved the best performance across datasets and prediction horizons, validating the synergistic role of each module in mitigating aliasing issues and enhancing prediction accuracy. Removing Mamba results in a significant performance drop for long prediction lengths, highlighting its importance in capturing nonlinear causal relationships and long-term trends in the temporal dimension. Removing Bi-Mamba leads to noticeable performance degradation, especially on the Weather dataset with complex inter-channel interactions, showing its role in modeling variable relationships. Comparatively, Mamba has a greater impact on long-sequence forecasting tasks, while Bi-Mamba is better at handling multivariable interactions. Together, they complement each other in unmixing across temporal and channel dimensions, improving the model's accuracy and robustness.

\begin{table}[!t]
	\centering
	\caption{Prediction performance of MTS-UNMixers under different history lengths (96, 192, 336, 720). The best performance is highlighted in bold.}
	\scalebox{0.85}{
		\begin{tabular}{c|c|cc|cc|cc|cc}
			\toprule
			\multicolumn{2}{c|}{History Length}& \multicolumn{2}{c|}{96} & \multicolumn{2}{c|}{192} & \multicolumn{2}{c|}{336} & \multicolumn{2}{c}{720} \\
			\midrule
			\multicolumn{2}{c|}{Metric}& MSE & MAE & MSE & MAE & MSE & MAE & MSE & MAE \\
			\midrule
			\multirow{4}{*}{ETTh1} & 96 & 0.368 & 0.388 & 0.361 & 0.379 & 0.359 & 0.377 & \textbf{0.355} & \textbf{0.371} \\
			& 192 & 0.427 & 0.419 & 0.424 & 0.413 & 0.419 & 0.409 & \textbf{0.416} & \textbf{0.412} \\
			& 336 & 0.443 & 0.433 & 0.441 & 0.431 & 0.438 & 0.431 & \textbf{0.432} & \textbf{0.428} \\
			& 720 & 0.454 & 0.429 & 0.448 & 0.427 & 0.444 & 0.414 & \textbf{0.439} & \textbf{0.409} \\
			\midrule
			\multirow{4}{*}{ETTm1} & 96 & {0.317} & {0.350} & 0.308 & 0.340 & 0.291 & 0.339 & \textbf{0.291} & \textbf{0.326} \\
			& 192 & 0.360 & 0.378 & 0.351 & 0.374 & \textbf{0.343} & 0.369 & {0.346} & \textbf{0.355} \\
			& 336 & 0.388 & 0.400 & 0.376 & 0.396 & \textbf{0.365} & 0.386 & {0.371} & \textbf{0.369} \\
			& 720 & 0.454 & 0.429 & 0.449 & 0.422 & {0.425} & {0.419} & \textbf{0.414} & \textbf{0.417} \\
			\midrule
			\multirow{4}{*}{Weather} & 96 & 0.158 & {0.195} & 0.158 & {0.191} & 0.153 & {0.189} & \textbf{0.149} & \textbf{0.183} \\
			& 192 & 0.206 & 0.242 & 0.200 & 0.240 & 0.194 & 0.239 & \textbf{0.193} & \textbf{0.235} \\
			& 336 & 0.254 & 0.286 & 0.254 & 0.279 & 0.248 & 0.274 & \textbf{0.243} & \textbf{0.271} \\
			& 720 & 0.339 & 0.336 & 0.335 & 0.329 & 0.328 & 0.328 & \textbf{0.316} & \textbf{0.324} \\
			\bottomrule
		\end{tabular}
	}
	\label{vary}
\end{table}
\begin{figure}[!t]
	\centering
	\captionsetup[subfigure]{labelformat=empty}
	\begin{minipage}{0.465\linewidth}
		\includegraphics[width=\textwidth]{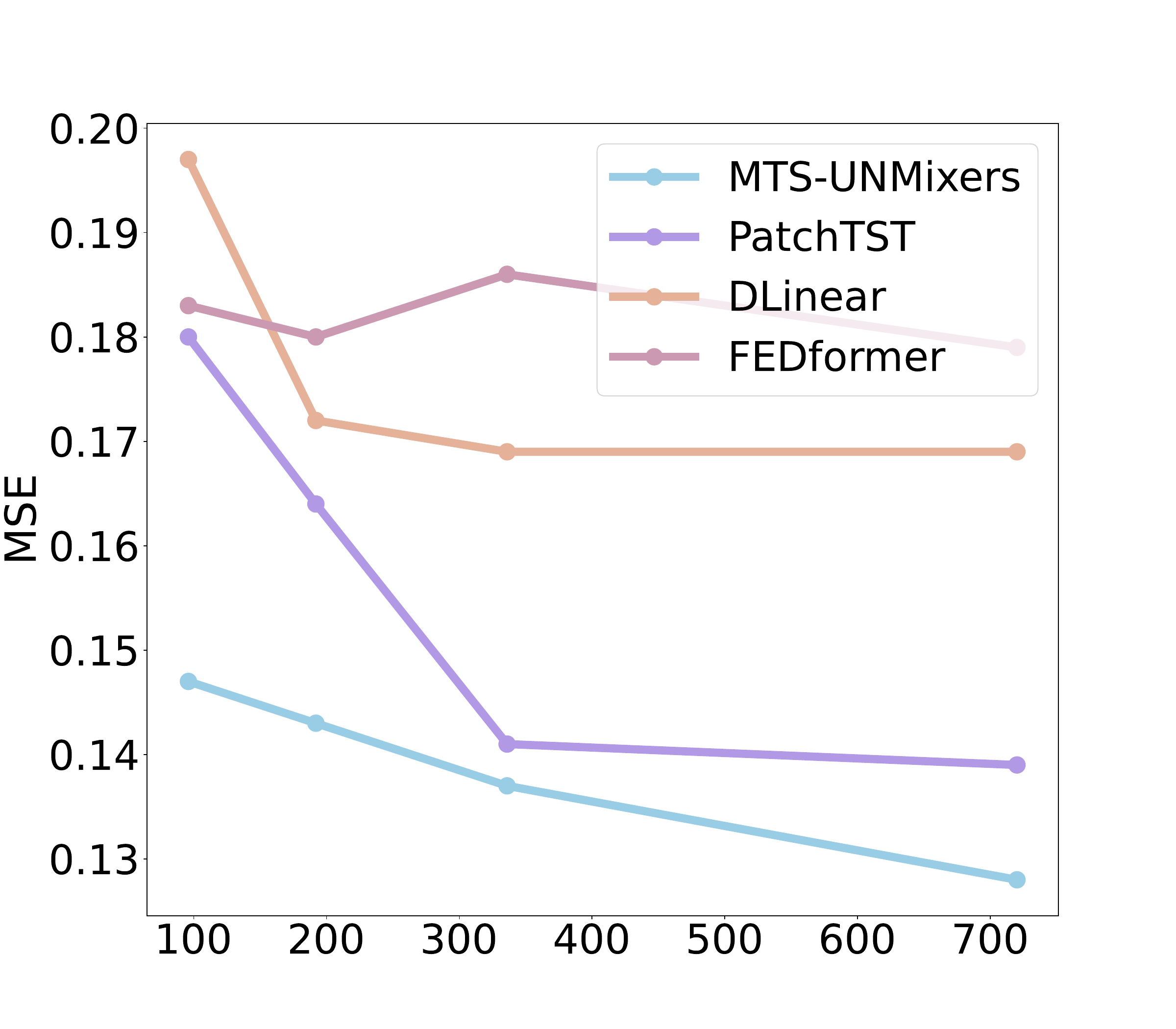}
		\subcaption{Electricity H  = 96}
	\end{minipage}
	\begin{minipage}{0.465\linewidth}
		\includegraphics[width=\textwidth]{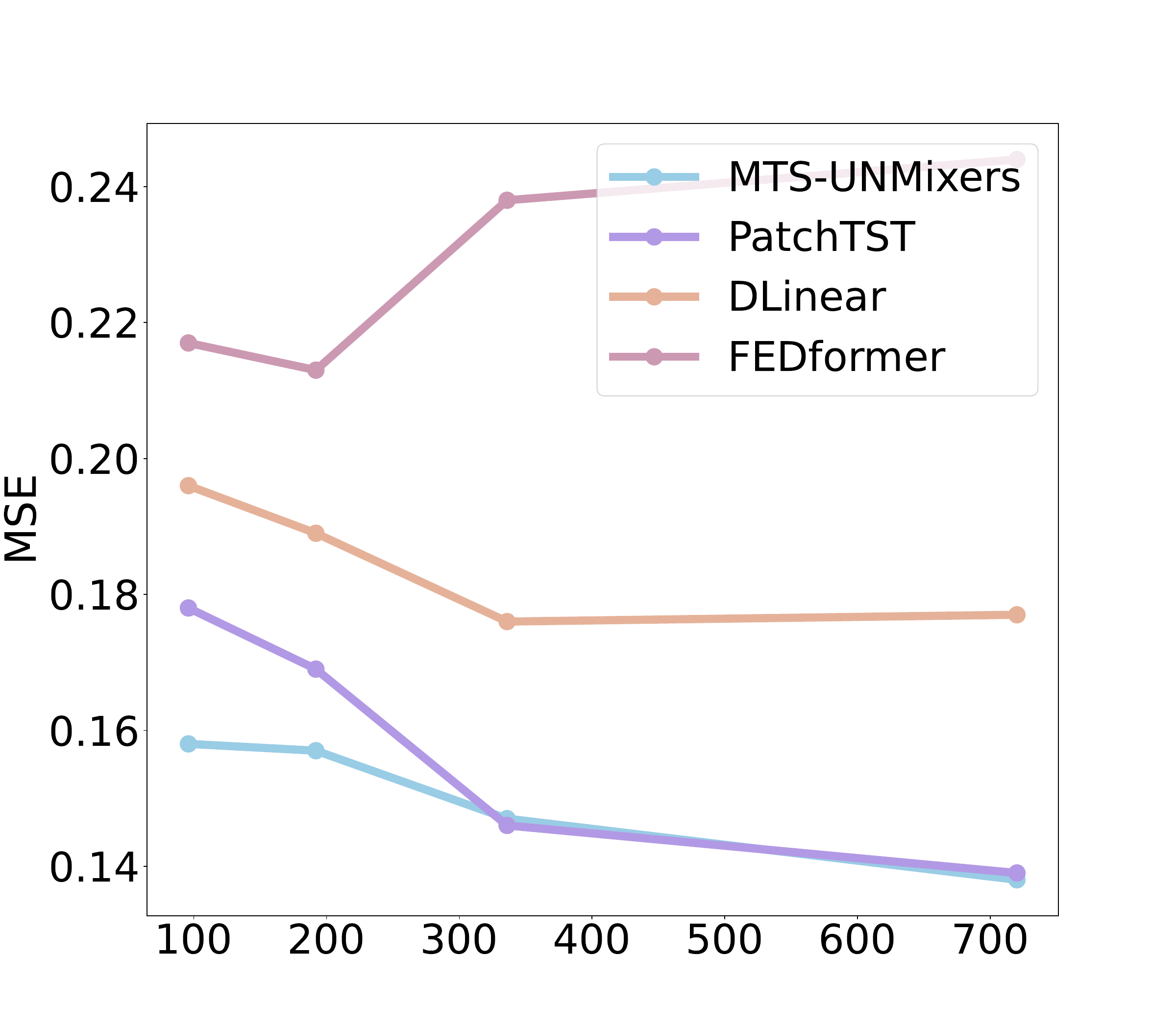}
		\subcaption{Weather H  = 96}
	\end{minipage}
	\caption{The predictive performance of different models for the future sequence \( H=96 \) under varying history lengths $\{96, 192, 336, 720\}$.}
	\label{lookback}
\end{figure}

\begin{table}[!b]
	\centering
	\caption{Model Efficiency Analysis.}
	\vspace{1pt}
	\centerline{(a) Running Time Efficiency Analysis (s/iter)}
	\vspace{3pt}
	\scalebox{0.65}{
		\begin{tabular}{c|c|c|c|c|c|c|c|c}
			\toprule
			 \multicolumn{2}{c|}{{\scriptsize Model}} & {\scriptsize Informer} & {\scriptsize Autoformer} & {\scriptsize FEDformer} & {\scriptsize PatchTST} & {\scriptsize TimesNet} & {\scriptsize TimeXer} & {\scriptsize MTS-UNMixers }\\
			\midrule
			\multirow{4}{*}{\makecell{Future Sequence \\ Length}}
			& 96  & 0.0078 & 0.0109 & 0.0859 & 0.0033 & 0.0428 & 0.0059 & 0.0054 \\
			& 192 & 0.0095 & 0.0111 & 0.0860 & 0.0033 & 0.0438 & 0.0050 & 0.0054 \\
			& 336 & 0.0098 & 0.0111 & 0.0864 & 0.0033 & 0.0528 & 0.0056 & 0.0054 \\
			& 720 & 0.0102 & 0.0111 & 0.0867 & 0.0033 & 0.0754 & 0.0056 & 0.0054 \\
			\midrule
			\multirow{4}{*}{\makecell{History Sequence \\ Length}} 
			& 96  & 0.0078 & 0.0109 & 0.0859 & 0.0036 & 0.0428 & 0.0059 & 0.0054 \\
			& 192 & 0.0079 & 0.0103 & 0.0777 & 0.0036 & 0.0408 & 0.0061 & 0.0054 \\
			& 336 & 0.0080 & 0.0104 & 0.0669 & 0.0036 & 0.0588 & 0.0061 & 0.0054 \\
			& 720 & 0.0082 & 0.0104 & 0.0669 & 0.0036 & 0.0791 & 0.0049 & 0.0054 \\
			\bottomrule
		\end{tabular}
	}
	\vspace{1pt}
	\vspace{1pt}
	\centerline{(b) Model Parameter Efficiency Analysis (M)}
	\vspace{1pt}
	\vspace{0.1pt}
	\scalebox{0.65}{
		\begin{tabular}{c|c|c|c|c|c|c|c|c}
			\toprule
			\multicolumn{2}{c|}{{\scriptsize Model}}&{\scriptsize Informer} & {\scriptsize Autoformer} & {\scriptsize FEDformer} & {\scriptsize PatchTST} & {\scriptsize TimesNet} & {\scriptsize TimeXer} & {\scriptsize MTS-UNMixers} \\
			\midrule
			\multirow{4}{*}{\makecell{Future Sequence \\ Length}}
			& 96 & 11.33 & 10.54 & 16.12 & 7.49 & 37.53 & 0.15 & 3.32 \\
			& 192 & 11.33 & 10.54 & 16.12 & 7.49 & 37.53 & 0.16 & 3.38 \\
			& 336 & 11.33 & 10.54 & 16.12 & 7.49 & 37.53 & 0.18 & 3.99 \\
			& 720 & 11.33 & 10.54 & 16.12 & 7.49 & 37.53 & 0.22 & 4.34 \\
			\midrule
			\multirow{4}{*}{\makecell{History Sequence \\ Length}} 
			&  96  & 11.33 & 10.54 & 16.12 & 7.49 & 37.53 & 0.15 & 3.32 \\
			& 192 & 11.33 & 10.54 & 16.12 & 7.49 & 37.53 & 0.16 & 4.27 \\
			& 336 & 11.33 & 10.54 & 16.12 & 7.49 & 37.53 & 0.18 & 5.76 \\
			& 720 & 11.33 & 10.54 & 16.12 & 7.49 & 37.53 & 0.22 & 10.18 \\
			\bottomrule
		\end{tabular}
	}
	\label{Eff1}
\end{table}
\subsection{Varying Lookback Window}
The results in Table \ref{vary} illustrate the impact of varying history lengths on the prediction performance of MTS-UNMixers across different prediction steps (96, 192, 336, and 720) on the ETTh1, ETTm1, and Weather datasets. Overall, longer history lengths contribute to improved prediction accuracy, particularly for longer prediction steps, highlighting the advantages of utilizing extended historical information. Theoretically, prediction performance should increase as the input history sequence length grows. 

For the ETTh1, ETTm1, and Weather datasets, extending the history length significantly enhances the prediction accuracy of MTS-UNMixers, with especially strong improvements for long prediction steps. For instance, increasing the history length from 96 to 720 across different datasets yields substantial improvements in both MSE and MAE, indicating that a longer history window helps the model better capture long-term dependencies and periodic features in time series data. In summary, the prediction performance of MTS-UNMixers consistently improves with extended history lengths, particularly in tasks involving longer prediction steps. These experimental results demonstrate that extending the historical sequence length effectively enhances the model's forecasting capability, underscoring the importance of capturing long-term dependencies in time series forecasting.

We also conducted a visual comparison with other models, as shown in Fig. \ref{lookback}. It can be seen that our results remain ahead, with a steady decrease in MSE and a gradual improvement in performance.

\begin{figure*}[!t]
	\centering
	\captionsetup[subfigure]{labelformat=empty}
	\begin{minipage}{0.4\linewidth}
		\vspace{3pt}
		\includegraphics[width=\textwidth]{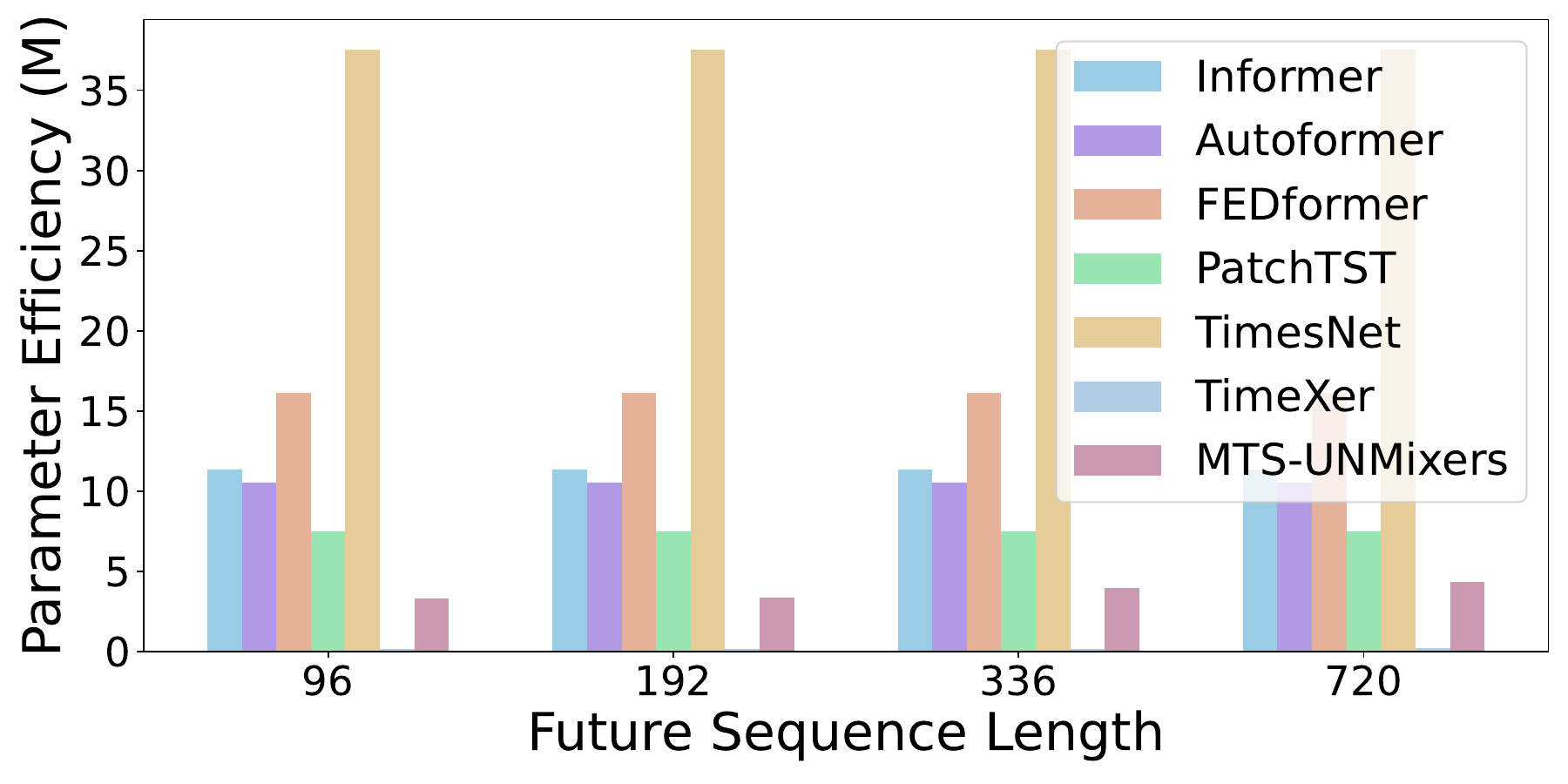}
		\vspace{3pt}
	\end{minipage}
	\begin{minipage}{0.4\linewidth}
		\vspace{3pt}
		\includegraphics[width=\textwidth]{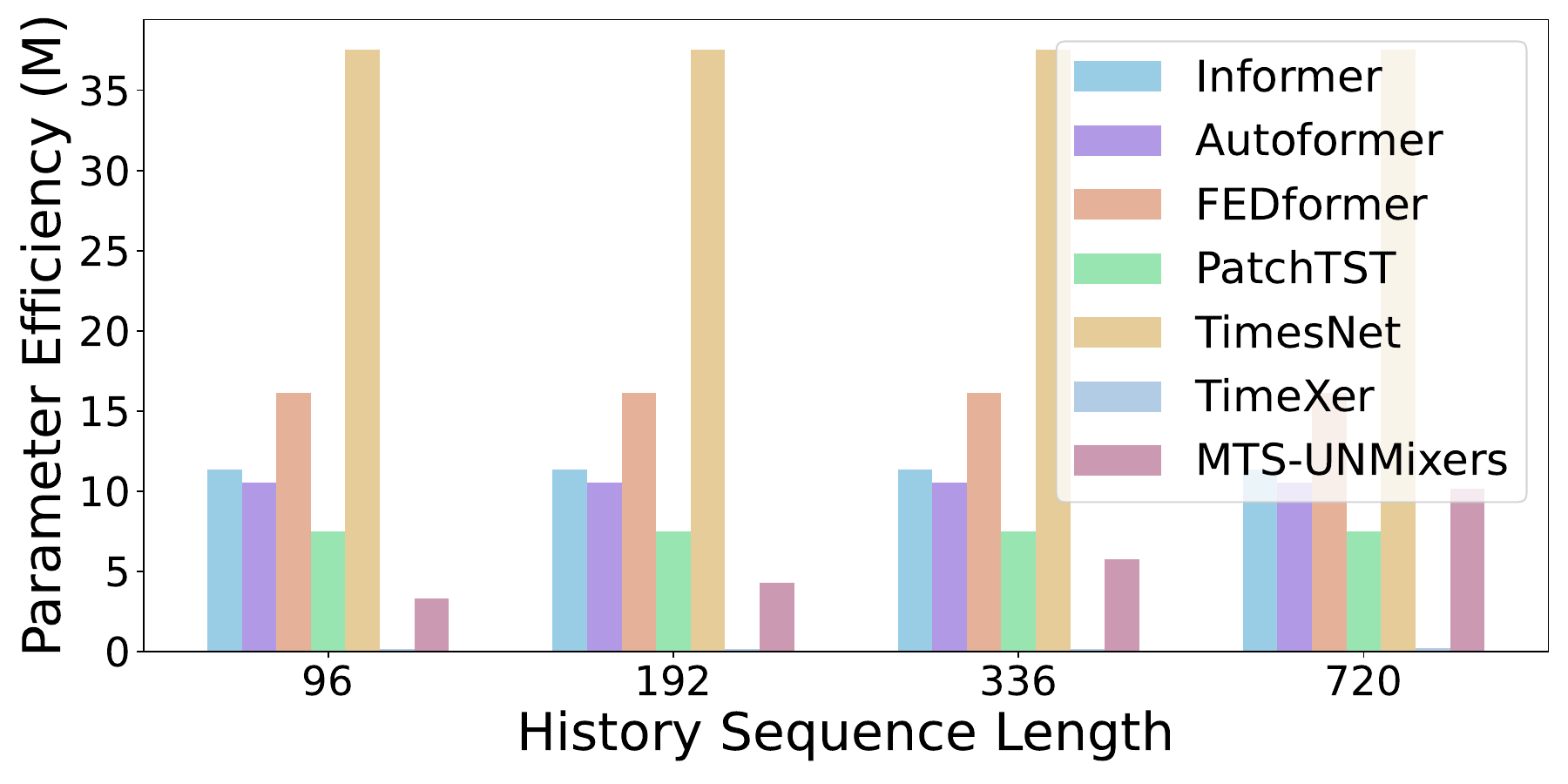}
		\vspace{3pt}
	\end{minipage}
	\begin{minipage}{0.4\linewidth}
		\vspace{3pt}
		\includegraphics[width=\textwidth]{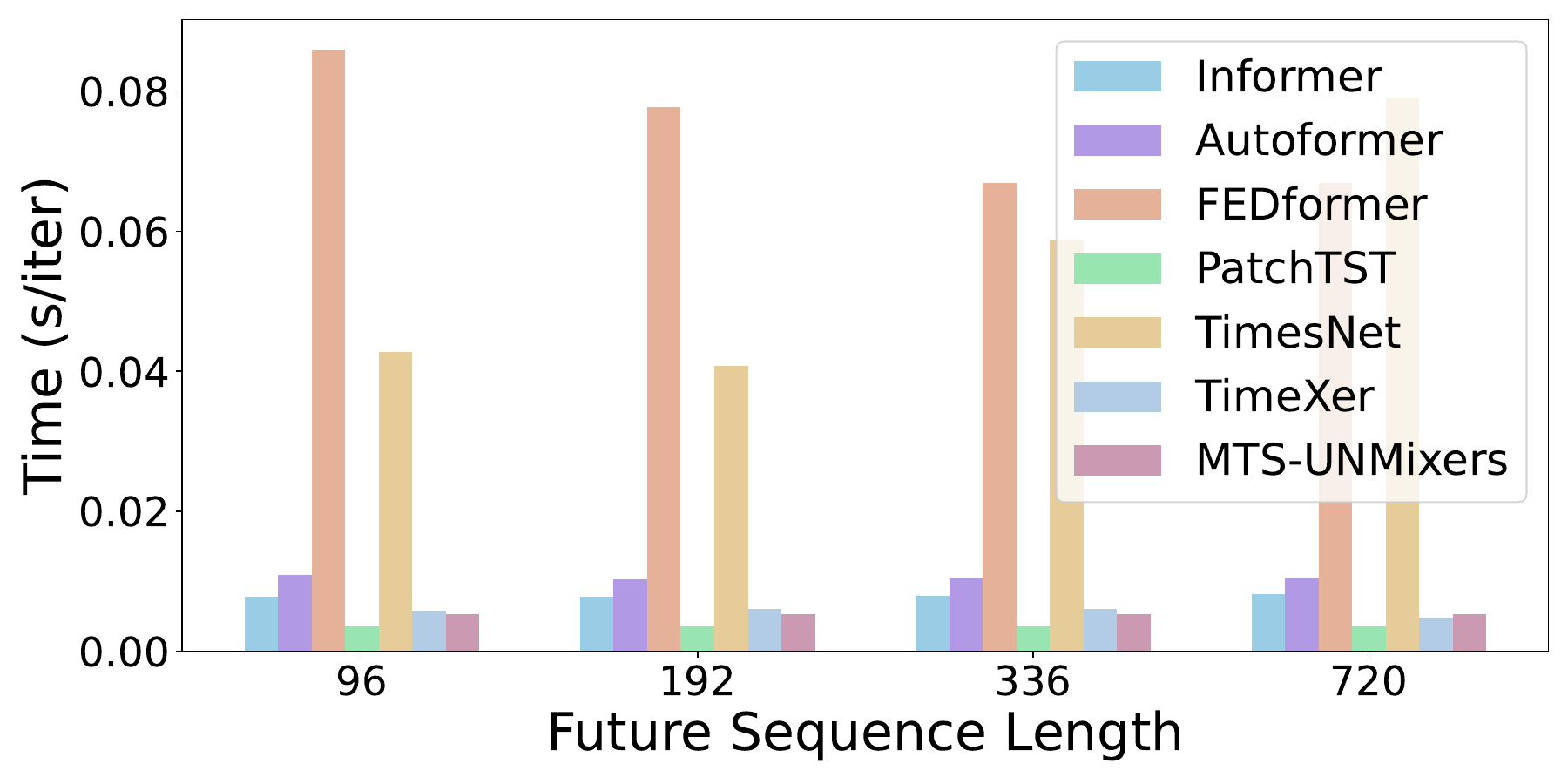}
		\vspace{3pt}
	\end{minipage}
	\begin{minipage}{0.4\linewidth}
		\vspace{3pt}
		\includegraphics[width=\textwidth]{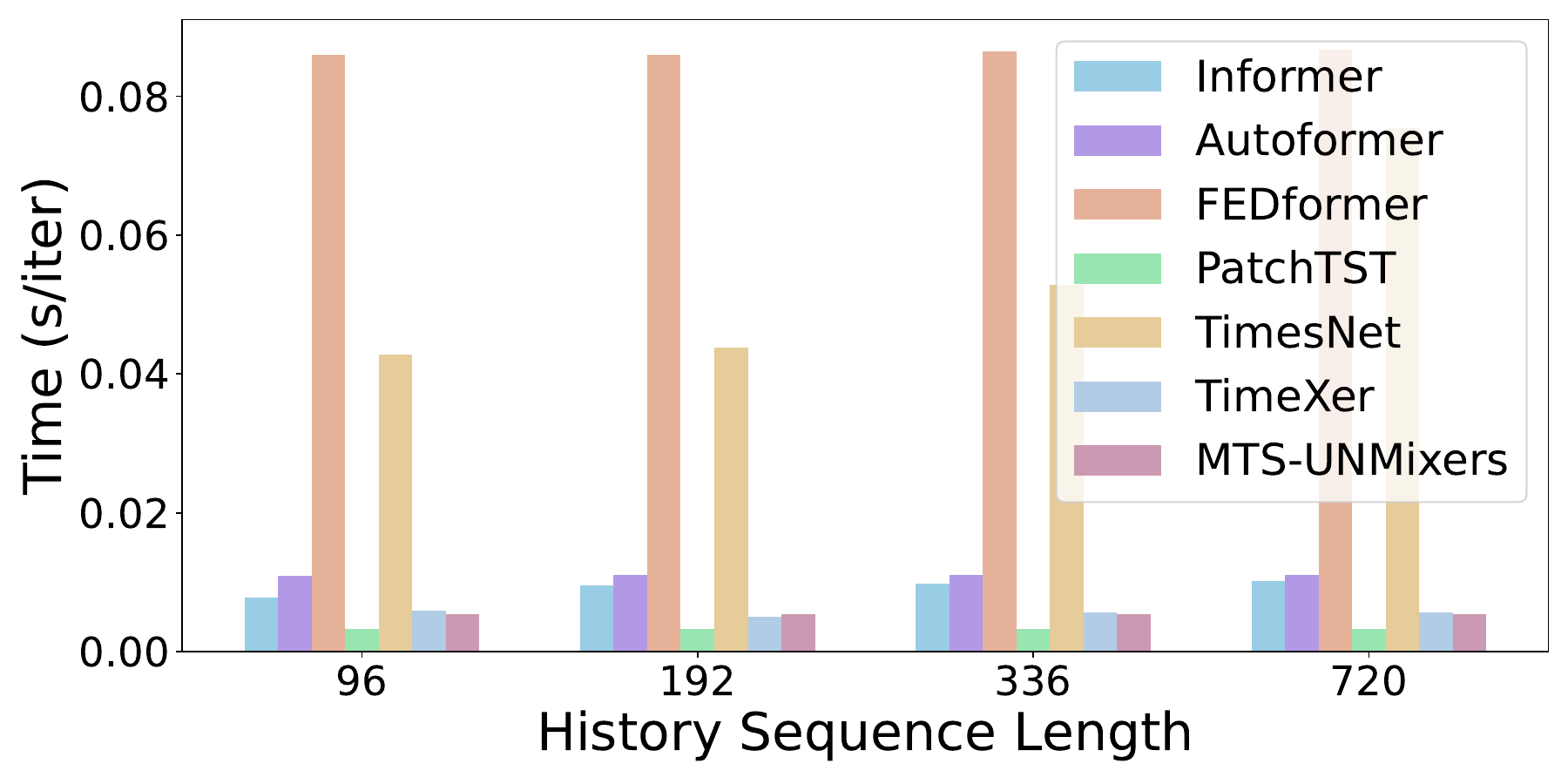}
		\vspace{3pt}
	\end{minipage}
	\caption{These four graphs show the speed and parameter efficiency of the model at different historical and future sequence lengths.}
	\label{eff2}
\end{figure*}	

\subsection{Model Efficiency Analysis}
To summarize the model performance and efficiency, we calculate relative performance rankings to compare the baselines. The ranking is based on the common models used across all five tasks: Informer, Autoformer, FEDformer, PatchTST, TimesNet, TimeXer, and our proposed MTS-UNMixers, totaling six models. We compare the models using three efficiency metrics under different input and output sequence lengths: the number of parameters (Params) and runtime (s/iter). All experiments were conducted at a unified level to ensure fairness. The results are presented in Table \ref{Eff1} and visualized in Figure 3. As shown in the Fig. \ref{eff2}, significant runtime differences exist between models with the same input and output sequence lengths. Although MTS-UNMixers does not outperform PatchTST, it still demonstrates competitive performance.

\section{Conclusion} \label{conclusion}
In conclusion, this paper presents MTS-UNMixers, a novel approach to time-series forecasting that leverages unmixing and sharing mechanisms within a Mamba-based network. By using Mamba blocks to separate channel coefficients and temporal basis signals, MTS-UNMixers captures complex inter-channel relationships and temporal dependencies with high precision. The integration of these unmixing mechanisms with a sharing phase enables efficient mapping of historical patterns to future predictions, effectively addressing challenges related to signal aliasing and redundancy. Extensive experiments on seven public datasets demonstrate that the model achieves superior performance compared to nine state-of-the-art baselines in various long-term forecasting tasks. The robust performance of MTS-UNMixers highlights its ability to effectively model intricate temporal dynamics and dependencies, offering significant advancements for multivariate time-series forecasting applications.

\vspace{12pt}

\end{document}